%% file: main.tex
\documentclass[10pt,twocolumn,letterpaper]{article}

\usepackage[pagenumbers]{cvpr} 

\input{preamble}
\usepackage{multirow}
\usepackage{array}
\graphicspath{{figures/}}

\definecolor{cvprblue}{rgb}{0.21,0.49,0.74}
\usepackage[pagebackref,breaklinks,colorlinks,allcolors=cvprblue]{hyperref}

\def\confName{CVPR}
\def\confYear{2027}

\newcommand{\method}{WeightCode\xspace}

\title{WeightCode: Installing New Capabilities into Neural Networks\\
via Weight-Space Programming}

\author{
Bo Li$^{1,2,*}$,
Taoyong Cui$^{3,*,\dagger}$,
Bob Zhang$^{1,\ddagger}$,
Mengran Li$^{4}$,\\
Chengyang Zhang$^{5}$,
Yang Zhang$^{2}$,
Pheng Ann Heng$^{3}$,
Wanli Ouyang$^{3}$,\\
Pengfei Li$^{6}$,
Luxia Zhang$^{7}$,
Zhenfei Yin$^{8,\ddagger}$,
Yingcheng Wu$^{9,\ddagger}$,
Ling Yang$^{7,\ddagger}$
\\[0.8em]
$^{1}$University of Macau \quad
$^{2}$National University of Singapore\\
$^{3}$The Chinese University of Hong Kong \quad
$^{4}$Sun Yat-sen University\\
$^{5}$Sichuan University \quad
$^{6}$Beijing Medical University\\
$^{7}$Peking University \quad
$^{8}$University of Oxford\\
$^{9}$Stanford University
\\[0.6em]
{\small
$^{*}$Equal contribution.
\quad
$^{\dagger}$Project lead.
\quad
$^{\ddagger}$Corresponding authors.
}
}

\begin{document}
\maketitle

\vspace{-0.8em}
\noindent\makebox[\textwidth][c]{%
    \small
    \textbf{Code:}~\href{https://github.com/Gen-Verse/WeightCode}{https://github.com/Gen-Verse/WeightCode}
}
\vspace{0.2em}

\input{fig/teaser}

\input{sec/0_abstract}
\input{sec/1_intro}
\input{sec/2_related}
\input{sec/3_method}
\input{sec/4_results}
\input{sec/6_discussion}

{
    \small
    \bibliographystyle{ieeenat_fullname_unsrt} 
    \bibliography{references}
}

\clearpage
\appendix

\setcounter{topnumber}{4}
\setcounter{dbltopnumber}{4}
\setcounter{bottomnumber}{2}
\setcounter{totalnumber}{6}

\renewcommand{\topfraction}{0.9}
\renewcommand{\dbltopfraction}{0.9}
\renewcommand{\bottomfraction}{0.6}
\renewcommand{\textfraction}{0.07}
\renewcommand{\floatpagefraction}{0.8}
\renewcommand{\dblfloatpagefraction}{0.8}

\input{sec/7_supplement}

\end{document}

%% file: preamble.tex
\usepackage{cuted} 

\usepackage{currfile} 

\usepackage{caption} 



%% file: fig/teaser.tex
\begin{strip}
\centering
\includegraphics[width=\linewidth]{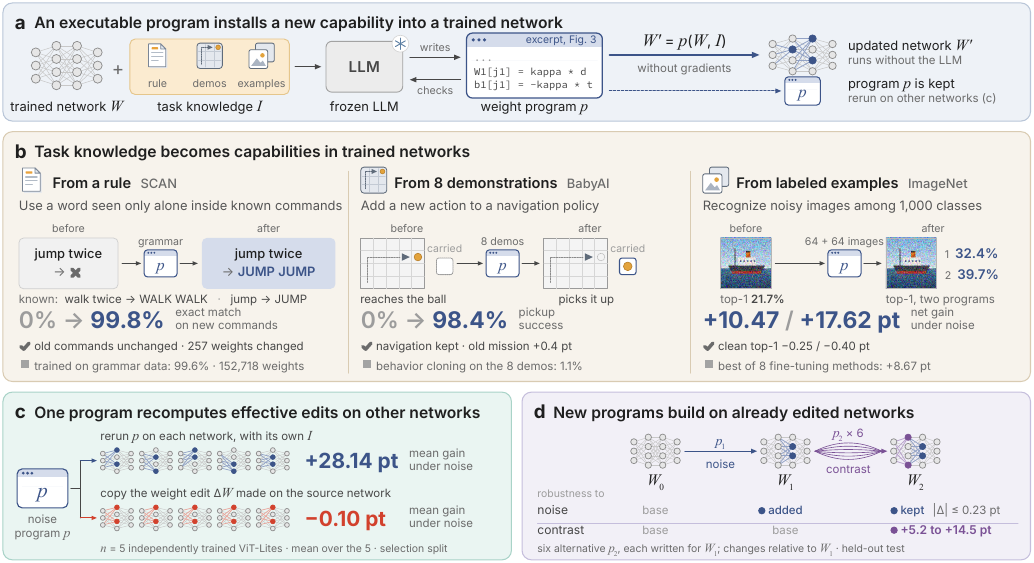}
\captionsetup{hypcap=false}
\captionof{figure}{
\textbf{\method installs, reuses, and extends capabilities through weight programming.}
(a)~A frozen LLM uses a trained network's weights $W$ and task knowledge $I$ to write an executable weight program $p$, which computes $W'=p(W,I)$ without gradients.
(b)~Weight programs turn rules, demonstrations, and labeled examples into new capabilities  across language, navigation, and visual recognition while preserving existing capabilities.
(c)~Rerunning the same program recomputes effective edits across independently trained networks, whereas copying its fixed weight edit does not reproduce the gain.
(d)~Programs written for an already edited network install a second capability while preserving the first.
}
\captionsetup{hypcap=true}
\label{fig:teaser}
\end{strip}

%% file: sec/0_abstract.tex
\begin{abstract}
Neural networks encode learned capabilities in their weights, yet extending those capabilities with explicit task knowledge remains a fundamental challenge.
Here we show that executable weight programs can construct and install new capabilities directly within trained neural networks.
\method uses a frozen language model to write executable weight programs that install new computations into trained neural networks using task knowledge and their existing weights, without gradient-based training. Once installed, the new capabilities run independently of the language model. Given a grammar or eight demonstrations, weight programs enable previously unseen command compositions (0\% to 99.8\% exact match) and a new pickup action (0\% to 98.4\% success), while preserving the original command and navigation capabilities. On ImageNet, four weight programs achieve net gains of 8.84–17.62 points while preserving clean accuracy, outperforming the best of eight fine-tuning methods under matched examples, checks, and retention constraints.
In a Vision Transformer, causal interventions show that installed noise-correction gates drive robustness gains, while their selective activation preserves clean accuracy.
Rerun on five independently trained Vision Transformers without further language model calls, ten weight programs retain 83–110\% of their source-network gains on average, whereas copying their fixed weight edits yields negligible gains.
Six programs written for an already edited network add 5.23–14.50 points of contrast robustness while retaining the previously installed noise gain.
Together, these results establish weight programming as a new route to acquiring and accumulating capabilities in trained neural networks through reusable executable programs.
\end{abstract}

%% file: sec/1_intro.tex
\section{Introduction}
\label{sec:intro}

Learning a capability during training and installing one into an already trained neural network are distinct problems. The latter requires translating task knowledge into new computations while preserving previously acquired capabilities. Unlike software, whose functionality can be extended through executable instructions, neural networks encode learned computations in distributed weights. Can executable programs directly construct new capabilities within these weights, and can subsequent programs build upon those already installed?

An executable weight program retains how a network-specific update is constructed, rather than only the numerical weight changes it produces. Fine-tuning~\cite{hu2021lora,li2018lwf}, model editing~\cite{meng2022rome,meng2022memit}, and test-time adaptation~\cite{schneider2020bn,wang2020tent} provide established ways to update trained networks. Meanwhile, large language models (LLMs) can generate and refine executable programs through evaluation feedback~\cite{romeraparedes2024funsearch,novikov2025alphaevolve} and have recently been shown to inspect and modify neural weights directly~\cite{takehi2026sparks}. Four fundamental questions define the scope of weight programming:
\begin{itemize}
\setlength{\itemsep}{1pt}
\setlength{\parsep}{0pt}
\item \emph{Installation.}
Can a weight program install a new capability into a trained network while preserving previously acquired capabilities?
\item \emph{Mechanism.}
What computation does a weight program install, and how does it produce the new capability?
\item \emph{Reuse.}
Can the same weight program be rerun without the LLM to construct effective edits in independently trained networks?
\item \emph{Succession.}
Can a new weight program extend an already edited network while preserving capabilities installed by earlier programs?
\end{itemize}

Here we introduce \method, which uses a frozen LLM to write executable weight programs that install new computations into trained neural networks (\cref{fig:teaser}a).
A weight program $p$ reads the network's weights $W$ and task knowledge $I$, such as a rule, demonstrations, or labeled examples, and constructs updated weights $W' = p(W,I)$.
Evaluation feedback selects programs that improve the new capability while preserving existing behavior. The updated network executes the installed computation within its original architecture, without the LLM at inference. The retained program can be rerun on compatible networks to construct network-specific edits. New programs can then extend networks already modified by earlier programs.

Across language, navigation, and visual recognition, weight programs transform rules, demonstrations, and labeled examples into new capabilities (\cref{fig:teaser}b, \cref{fig:sessions}). On SCAN, a program modifies 257 weights to raise exact match on previously unseen command compositions from 0.0\% to 99.8\%; on BabyAI, eight demonstrations enable a new pickup action with 98.4\% success.
On ImageNet, four weight programs achieve 8.84--17.62 points of net gain under corruption while preserving clean accuracy, outperforming eight fine-tuning methods
under matched examples, checks, and retention constraints.
Causal interventions reveal how installed noise-correction gates produce robustness gains while their selective activation preserves clean accuracy (\cref{fig:mech}).
Rerun on five independently trained Vision Transformers, ten weight programs retain 83--110\% of their source-network gains on average, whereas copying fixed weight edits yields negligible gains (\cref{fig:teaser}c, \cref{fig:reuse}).
Six second-round programs written for an already edited network add contrast robustness while retaining the previously installed noise robustness (\cref{fig:teaser}d, \cref{fig:stack}).
Together, these findings establish a new paradigm for extending trained neural networks: capabilities can be programmed into existing weights, reconstructed across
independently trained networks, and progressively expanded through successive weight programs.

%% file: sec/2_related.tex
\section{Related work}
\label{sec:related}

Existing approaches to modifying trained networks differ in how they construct, represent and reuse changes to model parameters.

\paragraph{Learning and retaining capabilities.}
Fine-tuning adds a capability by optimizing parameters on labeled data, in full or through parameter-efficient modules~\cite{houlsby2019adapters,hu2021lora}, and regularized transfer keeps the new parameters close to the old ones~\cite{li2018l2sp}.
Sequential training overwrites earlier capabilities~\cite{mccloskey1989catastrophic,french1999catastrophic}, and continual learning limits this forgetting by regularizing parameters~\cite{kirkpatrick2017ewc} or outputs~\cite{li2018lwf}, or by replaying stored examples~\cite{chaudhry2019tiny}.
Test-time adaptation adjusts a network to a new input condition from unlabeled inputs, by re-estimating normalization statistics~\cite{schneider2020bn}, minimizing prediction entropy~\cite{wang2020tent,niu2023sar} or training on a self-supervised task~\cite{sun2019ttt}.

\paragraph{Targeted and lifelong model editing.}
Model editing changes a trained network's behavior on target inputs while preserving it elsewhere: by gradient steps on a network trained to be editable~\cite{sinitsin2020editable}, by learned editor networks~\cite{mitchell2021mend}, or by closed-form rewrites of layers viewed as associative memories~\cite{bau2020rewriting,santurkar2021editing,meng2022rome,meng2022memit}.
Lifelong editing applies edits in sequence, for example by storing each one in a discrete key-value codebook attached to a layer~\cite{hartvigsen2023grace}.

\paragraph{Composing and transferring weight changes.}
Task arithmetic adds and subtracts the weight differences of fine-tuned models~\cite{ilharco2022taskarith}, and model merging resolves the interference between such differences when it combines them~\cite{yadav2023ties}. Both act on models that share a pretrained starting point, and merging independently trained networks first aligns their units by permutation~\cite{ainsworth2022rebasin}.
These methods operate on learned parameter differences, whose transfer and combination depend on the relationship between model parameterizations.

\paragraph{Programming neural computation and weight space.}
Language models have been used to discover executable algorithms, reward functions, policies, skills and training objectives~\cite{romeraparedes2024funsearch,novikov2025alphaevolve,ma2023eureka,liang2023cap,wang2023voyager,lu2024discopop}, and symbolic search and meta-learning discover optimizers~\cite{chen2023lion,andrychowicz2016l2l}.
Separately, work on hypernetworks, weight-space learning and program-to-network compilation has established several ways to construct or analyze neural computation through weights~\cite{ha2016hypernetworks,navon2023dws,zhou2023nfn,schurholt2021hyperrep,weiss2021rasp,lindner2023tracr,friedman2023ltp}.

\paragraph{Weight-space programming.}
In concurrent work, \citet{takehi2026sparks} show that a frontier LLM can read the weights of small networks, write networks from zero weights and edit trained networks, and \citet{li2026weightloop} have an LLM write an editor that refits the output weights or coefficients of trained predictors from measured feedback.
These studies show that language models can inspect and modify neural weights. \method shows that the programs that construct such updates, retained after the update, can be rerun on independently trained networks and extended by subsequent programs, so that the construction procedure itself becomes a reusable object.

%% file: sec/3_method.tex
\section{\method}
\label{sec:method}

\method separates the construction of an update from its execution: a frozen LLM writes an executable program, and the program computes network-specific weight changes (\cref{fig:teaser}a).

\subsection{Installing a capability with a program}
\label{sec:method_install}

\paragraph{Weight programs.}
A weight program $p$ is code that takes a trained network's weights $W$ and task knowledge $I$, what is known about a new capability, and returns new weights:
\begin{equation}
W' = p(W, I), \qquad \Delta W = p(W, I) - W.
\end{equation}
$\Delta W$ is the \emph{weight edit} of one run. It depends on the network the program reads as well as on the program, so the same program run on another network computes another edit.
A program may read every weight, run the network's forward pass on inputs and read $I$.
The updated network keeps its architecture and forward code, with only its weights changed, and runs on its own at inference.
The experiments impose additional implementation constraints, including sandboxed execution and a maximum edit size of 20\% of the network's weights (\cref{sec:supp_impl}).

\paragraph{What a program writes.}
In \method, the form of the update is chosen for each capability and each network.
The LLM proposes candidate computations and weight modifications from the network's weights, its activations and the task knowledge, and evaluation feedback determines which candidate is kept.
The programs reported here connect the new capability to computation the network already performs (\cref{tab:what}). The Gaussian-noise program for the ViT-Lite, for example, measures how noise shifts the patch tokens, finds MLP units that are nearly idle on clean images, and rewrites two of them in each of three blocks as gates that open when a token moves along the noise direction and then subtract the measured shift (\cref{fig:mech}a).
Other programs reuse the network's own representations, such as the embeddings of known primitives on SCAN or of a known instruction in the BabyAI policy.

\paragraph{Writing and checking.}
The LLM writes the program in one session, in a Python kernel that holds $W$, the forward pass and $I$.
Program construction alternates between inspecting the network, writing candidates and evaluating them.
A check runs a candidate on the network and returns two counts on a check split that $I$ does not contain: $S_\text{new}$, the inputs of the new condition that the candidate fixes minus those it breaks, and $S_\text{old}$, the same count on old inputs.
A session has 8 checks.

\paragraph{Keeping existing behavior.}
Preservation is evaluated on task-specific reference conditions: clean-image accuracy for the image classifiers, the previously supported commands or mission for the sequence and navigation tasks, and the previously installed conditions for successive updates. Keeping these is part of how a program is chosen.
The kept program is the checked candidate with the highest $S_\text{new} + S_\text{old}$ among those with $S_\text{old} \ge -20$, a loss of at most 0.5\% of 4,000 clean check images (the \emph{retention rule}). On SCAN the bound is 5 of 1,000 old commands and on BabyAI 3 of 500 old episodes. If no candidate qualifies, the network is left unchanged.
The bound is applied on the check split, and the reference conditions are measured on held-out inputs (\cref{sec:protocol}).

\paragraph{Feedback without gradients.}
Weight programs construct updates through direct computations on weights, activations and forward passes, namely weight and activation statistics, SVD and PCA, label-free closed-form fits and single-unit ablations, rather than by optimizing an objective on the network's outputs.
Labeled feedback evaluates and selects candidate programs rather than fitting their values: a single-round session scores at most 8 candidate edits against the labeled examples and receives 8 checks, as each fine-tuning method tries 8 configurations and receives the same checks. Label-scored work that changes no weight, and selection inside a program's own run, are counted separately from this cap (\cref{sec:supp_impl}).
An automatic audit and a reading of the kept programs check these rules; three exploratory sessions that use finite differences are reported separately (\cref{sec:supp_fd}), and \cref{sec:supp_impl} lists what the protocol admits.

\subsection{Rerunning and extending programs}
\label{sec:method_reuse}

\paragraph{Programs as reusable construction procedures.}
An executable weight program retains the procedure that constructed an update: how the network is measured, how units or directions are selected and what is written into them.
Run on another compatible network, the same program remeasures that network and constructs a new edit in its weight coordinates.
This matters for independently trained networks, which may implement related functions in different coordinates~\cite{ainsworth2022rebasin}, so that a numerical edit computed on one network need not apply to another.

\paragraph{Copying, rerunning and successive updates.}
Three operations act on a program $p_1$ kept for a network $W_0$ and task knowledge $I_1$, which produces $W_1 = p_1(W_0, I_1)$.
\emph{Copying} adds the edit $W_1 - W_0$ unchanged to another network.
\emph{Rerunning} executes the code of $p_1$ unchanged and without another LLM session, while the program inspects the target network and uses that network's own task examples, so each rerun computes an edit from that network's weights and data.
A \emph{successive update} asks the LLM for a new program $p_2$ for the edited network $W_1$, which produces $W_2 = p_2(W_1, I_2)$. This new program is constructed with the edited network, the code of $p_1$, labeled examples of the new condition and checks that also report every earlier condition, under the retention rule for each, so that the update can account for the computations installed earlier.
We rerun programs on independently trained networks of the architecture they were written for, and write second programs for edited networks.

\subsection{Experimental protocol}
\label{sec:protocol}

\paragraph{Networks.}
Programs are written by Claude Opus 5.5 at maximum reasoning effort.
The image networks are a Vision Transformer (ViT)~\cite{dosovitskiy2020vit}, ViT-Lite-7/4~\cite{hassani2021cct} (3.7M weights, 89.35\% clean accuracy), and a ResNet-20~\cite{he2016resnet} (0.27M weights, 91.28\%), both trained on 40,000 CIFAR-10 training images, and torchvision's ImageNet~\cite{russakovsky2014imagenet} ResNet-18 in two forms: unchanged on its 1,000 classes (11.7M weights, 69.70\% on our ImageNet test images) and fine-tuned to CIFAR-10 (11.2M weights, 95.78\%).
The other networks are a BabyAI policy trained by imitation to go to a named object and a Transformer per SCAN split~\cite{lake2018scan,loula2018rearranging}.
Five further networks per CIFAR-10 architecture, trained from other seeds, serve reuse.

\paragraph{Conditions.}
Corruptions are the CIFAR-10-C set at severity 5~\cite{hendrycks2019cifarc} and, on ImageNet, the ImageNet-C functions at the highest severity that costs at least 15 points and keeps at least 17\% of the network's accuracy above chance: 3 for Gaussian noise and 4 for contrast reduction.
A condition is run on a network when it costs the network at least 15 points of accuracy, which on the ViT admits Gaussian noise and contrast reduction.
\Cref{tab:protocol} summarizes what each task gives the LLM and the references, and \cref{sec:supp_impl} gives training, data and sandbox details.

\paragraph{Task knowledge.}
$I$ comes at several levels.
\emph{Labeled examples}: 64 labeled images of the new condition (16 that the network misclassifies and 48 drawn at random, with the network's predictions) and 64 labeled clean images. For the BabyAI policy, 8 demonstrations of the new mission (each step's observation, instruction and expert action) and 64 old steps.
\emph{Condition labels withheld}: the LLM receives the same condition images and the network's predictions but not their labels, and no statement of how they were chosen. The image pool and the selection are unchanged, including the use of labels to pick the 16 misclassified images, and the clean images and the checks keep their labels.
\emph{A rule}: SCAN's published interpretation grammar with 64 old commands and, on the add-primitive split, the training row of the new primitive. For the ViT, the formula that generates the corruption and 64 labeled clean images, with no image of the condition.
\emph{A capability's name}: one sentence asking for robustness to additive Gaussian pixel noise, with no image, label or prediction, given with the 8 checks or with none, in which case the one program the LLM declares is scored once after the session. This is the only level at which the LLM is told the condition.

\paragraph{References and criterion.}
Every fine-tuning comparison fits the same labels by gradient.
It has eight methods: full, last layer, LayerNorm only (BatchNorm on the ResNet-18s), LoRA~\cite{hu2021lora} with rank 4 and 16, full without the clean images, full with LwF~\cite{li2018lwf} and full with L2-SP~\cite{li2018l2sp}.
Comparisons match the task examples, the number of external checks and the retention rule: each method sees the same labeled images (the method without clean images only the 64 of the condition), tries 8 configurations, receives the same 8 checks and is kept by the same rule. The methods differ in how they construct updates, including the pretrained prior of the LLM that writes each program. We also report each method's best configuration on the check split without the clean-loss bound.
Eleven fixed references per image network, label fits and label-free rules including Tent~\cite{wang2020tent}, BN adaptation~\cite{schneider2020bn} on the ResNets and SAR~\cite{niu2023sar} on the ViT, are kept by the same rule.
Before the sessions on the ViT and the ResNet-18s we fixed a performance criterion for the comparative evaluation, the \emph{success criterion}: a gain of at least 10 points (on ImageNet, 15\% of the accuracy the condition costs: 7.0 and 8.5 points), a clean change of at least $-$1.0 and a net gain at least one point above the strongest reference that obeys the program limits (for most conditions, gradient training of a fixed 19\% of each tensor's entries) given the labeled examples. Sessions given less information on the ViT are read against the same bar.

\paragraph{Splits.}
The held-out images form three disjoint splits per condition: the condition images of $I$ are drawn from an evidence split (2,000 images), the checks use a check split (4,000) and analyses are read on a selection split (4,000). The clean images of $I$ are one fixed draw from the network's training images.
On ImageNet the 50,000 validation images are divided by image: 12,000 form these three splits and the pool of clean images for $I$, and the other 38,000 are the test set.
The selection split plays no part in any session, check, retention rule or fine-tuning configuration choice.
The tables report the held-out test sets: the official CIFAR-10 and CIFAR-10-C test files, the 38,000 ImageNet test images, the third of SCAN's test commands kept for the final evaluation, and held-out BabyAI episodes. Each kept network was read on them once, after every selection was final.

\paragraph{Metrics and reporting.}
\emph{Gain} is the accuracy change on the condition, \emph{clean change} the accuracy change on clean images, and \emph{net gain} their sum.
The mechanism analyses (\cref{sec:mechanism}) and the copied edits (\cref{sec:reuse}) are read on the selection split, and other numbers that exist only on the selection split are marked where they appear.
Each program is one LLM session on one trained network. On the ViT we ran ten sessions, three per condition on the seed-100 network and one per condition on networks trained from seeds 101 and 102, and report all of them (\cref{tab:sessions}). On the ImageNet ResNet-18 we ran two sessions per condition and report both (\cref{tab:imagenet}). \Cref{tab:main} shows the first session on each network.

%% file: sec/4_results.tex
\section{Results}
\label{sec:results}


\subsection{Weight programs turn task knowledge into new capabilities and keep existing behavior}
\label{sec:examples}
\label{sec:info}
\label{sec:rule}
\label{sec:unlabelled}

A rule becomes a new primitive inside existing computation.
SCAN's add-primitive split~\cite{lake2018scan} trains a Transformer on every command except those that compose the primitive \emph{jump}, which it sees only alone, and the network executes none of these compositions (\cref{fig:sessions}a, \cref{tab:behaviour}).
Given SCAN's interpretation grammar, 64 old commands and the \emph{jump} row, the program raises exact match on them from 0.0\% to 99.8\% by changing 257 weights, and the old commands do not change.
It moves the embedding of \emph{jump} to the centroid of those of \emph{walk}, \emph{look} and \emph{run} and adds an output row for its action, so the composition the network already computes now carries the new primitive.
A rule-to-data reference, trained on commands generated from the grammar, reaches 99.6\% with 152,718 changed weights, about 594 times as many, and fine-tuning on the given rows keeps at most +1.21 points of net gain (selection split).
A compact rule can therefore be installed by reusing the Transformer's existing compositional computation, with 257 modified weights.

The same principle extends to a different compositional setting.
On the around-right split~\cite{loula2018rearranging}, where \emph{around} was trained only with \emph{left}, the program writes the rule directly into 25,056 weights, as rank-one edits of attention values along the network's own right-minus-left direction, and raises exact match on commands with \emph{around right} from 17.6\% to 89.6\% at $-$0.3 points on the old commands, where fine-tuning on the given rows keeps at most +2.61 points of net gain (selection split; the rule-to-data reference is in \cref{fig:sessions}a and \cref{tab:behaviour}).

Eight demonstrations become a new action.
A BabyAI policy~\cite{chevalierboisvert2018babyai} trained by imitation to go to a named object never uses the pickup action.
Given 8 demonstrations of the mission \emph{pick up} and 64 old steps, the program raises pickup success from 0.0\% to 98.4\%, and the old mission moves by +0.4 points.
The installed behavior generalizes beyond the demonstrations: success remains 98.9\% on episodes whose object color and type never appear in the demonstrations (selection split).
Behavior cloning on the same demonstrations reaches at most 1.1\% while keeping the old mission (\cref{fig:sessions}a).
The program thus adds pickup to the navigation the policy already performs while retaining its original mission (\cref{sec:mechanism}).

With the same examples, programs reach larger gains than gradient training on ImageNet.
On torchvision's ResNet-18 unchanged on its 1,000 ImageNet classes, every method receives the same 64 labeled images of a corruption and 64 labeled clean images, the same 8 checks and the same retention rule (\cref{fig:sessions}b, \cref{tab:main}).
The two programs written for Gaussian noise reach net gains of +10.47 and +17.62 points and the two written for contrast reduction +8.84 and +14.26, against +8.67 and +3.38 for the best of eight fine-tuning methods and +8.84 and +6.00 for the best fixed reference (\cref{tab:imagenet}).
The four programs also keep clean accuracy within 0.89 points of the original network on ImageNetV2~\cite{recht2019imagenetv2}, a test set collected separately following ImageNet's procedure.

The result holds across architectures on CIFAR-10: the first programs reach net gains of +32.79 and +16.51 on the ViT-Lite, against +22.78 and +7.89 for the best fine-tuning, and +35.23 and +21.28 on the ResNet-18 fine-tuned to CIFAR-10, 3.7 and 2.1 times the best fine-tuning.
The resulting edits vary with the network and the computation, from 3,846 modified weights for the ViT-Lite noise gate to 2,152,928 for the ImageNet noise correction (0.10\% to 18\% of each network).

Independently written programs also exceed matched fine-tuning: on three ViT-Lites, eight of ten programs written in separate sessions do so on their own network, with net gains from +6.61 to +32.79 (\cref{fig:sessions}c, \cref{tab:sessions}).

\input{fig/fig_sessions}

The program's lead is largest where existing behavior must be kept and labeled data are scarce.
On the ViT-Lite under Gaussian noise, full fine-tuning first matches the program's net gain within the retention rule at 4,096 labeled corrupted images and as many clean ones, compared with 64 of each for the program (selection split, \cref{fig:f2}).
At clean-loss bounds of 0.5 and 1 point on the check split, the program has the highest net gain of every method on all four CIFAR-10 instances, by +2.7 to +25.5 points; \cref{sec:supp_ft} and \cref{fig:candidates} chart the full trade-off between gain and retention.

Programs can be written from the condition images without their labels.
On the ViT-Lite under Gaussian noise, with the labels of the 64 noisy images withheld from the LLM and the network's own predictions in their place, the program reaches a net gain of +34.43 at a clean change of $-$0.04, compared with +32.79 for the session given all labels (\cref{tab:main}).
The images are selected as in the labeled setting (16 that the network misclassifies), and the clean images and the 8 checks are labeled.
Given only the capability's name, programs reach +9.43 and +15.10 on the selection split (\cref{fig:ladder}).
Direct access to the condition labels is thus not required in this setting: the network's predictions, statistics of the images and labeled check feedback suffice to construct an effective correction.

\input{fig/fig_mech}

\subsection{Weight programs install computations that can be tested part by part}
\label{sec:mechanism}

Weight programs produce identifiable internal computations whose causal contributions to the new behavior can be tested directly (\cref{fig:mech}; \cref{tab:what} describes every program and marks those tested by intervention; selection split).

The noise program writes a conditional correction into existing units.
The installed correction detects a shift in the network's internal representations and subtracts a measured offset selectively from the affected tokens.
By examining the supplied images and the network's activations, the LLM finds that the corruption mainly shifts the patch tokens in the early residual stream, and builds the correction on that shift; untold, it also infers additive white Gaussian pixel noise of standard deviation about 25/255 (the generating value is 0.10).
The program measures this shift on synthetic noisy copies of the clean images, frees two little-used MLP units in block 2 by folding their mean output into the bias, and rewrites them as a ramp gate,
\begin{equation}
g(z) = \mathrm{GELU}\big(k(d^\top z - t)\big) - \mathrm{GELU}\big(k(d^\top z - t) - K\big),
\end{equation}
which subtracts the measured offset only from tokens whose projection $d^\top z$ on a whitened noise direction lies above the clean range (\cref{fig:mech}a). Two gates of the same form act on the class token in blocks 4 and 6.

The installed gates carry the gain.
Removing the block-2 gate lowers the net gain from +31.75 to +10.05, and removing all three gates returns the network to its original accuracy (\cref{fig:mech}c).

Selective activation keeps clean accuracy.
The block-2 gate opens on 99.5\% of noisy and 0.5\% of clean patch tokens and cuts the mean noisy-minus-clean offset from 5.29 to 0.50, while the program moves clean patch tokens by 0.05 on average against 5.02 on noisy ones (\cref{fig:mech}b).
Holding the gates open lowers the net gain to +19.43 and clean accuracy by 5.03 points, 15.5 on the worst class.
Gate activation also tracks where the correction helps: on 25 corrupted versions of the test set, the program helps on the whole noise family and on frost, where the gate opens on 27\% to 100\% of patch tokens, and changes no other set by more than 0.27 points on the test sets (\cref{fig:range}).

The gated computation matters more to the gain than the identity of the units that carry it (\cref{fig:mech}c).
The gated structure keeps the gain when its parts are replaced: a gate built from the plain mean difference of the noisy and clean examples keeps +31.13, and the same gate in two units drawn at random from the block gives +31.86 (\cref{tab:controls}).
Gradient training of exactly the entries the noise and contrast programs change, from their original values with the fine-tuning grid and the same rule, reaches +3.60 and +0.72, and from the programs' values it keeps +33.35 and +18.00: the program's specific weight configuration, rather than the choice of edited entries alone, accounts for the gain.
Network-specific information shapes the program: with access to the trained weights and outputs, the programs reach +31.75 and +17.98, against +10.78 and +11.18 when the same LLM receives the parameter names and shapes, the forward code, the examples and the checks, but no weight value or network output (one session per condition).

Different capabilities receive different computations.
Unlike the noise program, the contrast program combines an input-wide adjustment of the patch-embedding filters with a condition-selective correction (\cref{fig:mech}d). It raises the zero-mean part of the filters by a factor of 1.45, a global gain on the first layer, and adds a gated rank-6 correction in eight freed units of block 2, whose gate, on its own, opens on 99.7\% of low-contrast and 0.3\% of clean patch tokens.
On the original network the gain alone adds +10.95 points and the correction alone +9.63, at clean changes of 0.00 and +0.30, and together they reach +17.93: each part adds 7.0 and 8.3 points on top of the other. The noise computation contains no such first-layer gain; run as a template on contrast, the noise gate reaches +5.40 (\cref{tab:controls}).
The two components also differ in reach: on the ViT-Lite and the CIFAR-10 ResNet-18 alike, the input-wide gain accounts for the changes on other corruptions, the gains under blur and fog and the losses under noise, whereas the gated correction acts selectively on contrast reduction (\cref{fig:range}).
Across architectures and target conditions, programs construct different forms of computation, including gates in little-used units, corrections of activation shifts, input-layer rescaling and convolutional filtering (\cref{tab:what}). Across the ten ViT-Lite programs, those that meet the criterion repurpose little-used capacity into a condition-specific computation: gated corrections in freed MLP units, or a nearly unused attention head rewritten to subtract the image's mean color.

The BabyAI program writes a stated rule into the policy, a complementary form of mechanistic evidence.
It copies the embeddings of \emph{go} and \emph{to} into \emph{pick} and \emph{up}, builds from dead units an exact detector for a faced cell of the named color and type that raises the pickup action, and adds a detector that stops the policy from pressing forward into a blocking object.
An explicit rule reconstructed from these three parts predicts the edited policy's success or failure on all 2,000 evaluated episodes, 1,000 with the new mission and 1,000 with the old. The weights carry the stated rule, and the policy runs it with no outside code at the base network's inference cost.

\input{fig/fig_reuse}

\subsection{Executable programs recompute effective edits across independently trained networks}
\label{sec:reuse}

When the same program is executed on five independently trained ViT-Lites, using each network's own 64 labeled examples, it constructs a network-specific edit without another LLM session.
Rerun this way, the Gaussian-noise program gains +27.55 to +30.54 points (mean +28.57) at clean changes of $-$0.19 to +0.07, and the contrast program +16.04.
All ten programs carry their gains over: on the five networks each was not written for, each keeps 83\% to 110\% of its own gain on the held-out test sets, and clean accuracy stays within 1 point on 48 of the 50 transfers (\cref{fig:reuse}c).
Reused the same way, the best fine-tuning recipe reaches +20.79 and +8.09 (\cref{tab:reuse}).

Across independently trained ViT-Lites, rerunning a program transfers its effect more reliably than copying the source network's numerical edit (\cref{fig:reuse}a,b).
Added unchanged to the other ViT-Lites, the edits of the ten programs gain $-$1.42 to +0.01 points on average, 7.6 to 28.2 below rerunning the programs (selection split).
A rerun recomputes where and how to write: on one network the noise program gates blocks 2, 3 and 6 instead of 2, 4 and 6, with thresholds measured on that network (\cref{fig:reuse}a, \cref{sec:supp_reuse}).
Matching units does not recover this: aligned unit by unit to the original by activation matching on 2,048 clean images, the copied edits of the first program of each condition gain $-$0.01 and +4.17, whereas each rerun reconstructs its edit from the target network's own computation.
Copied edits transfer where networks share coordinates: when ResNet-18s share a pretrained initialization, they keep 96\% and 87\% of the rerun gain (selection split, \cref{fig:reuse_r18}), while a program transfers across independently trained networks as well.

\input{fig/fig_stack}

\subsection{Successive programs extend edited networks and keep earlier gains}
\label{sec:stack}

Executable weight programs install capabilities in succession.
On the ViT-Lite, after the Gaussian-noise program, six contrast programs written in separate sessions for the edited network add +5.23 to +14.50 points of contrast robustness on the held-out test sets, while the installed noise robustness moves by $-$0.17 to +0.23 and clean accuracy stays within 1.47 points of the original network (\cref{fig:stack}a,c, \cref{tab:rounds}); four exceed the 10-point threshold for the additional capability fixed before the sessions.
In the reverse order, all six noise programs written for the contrast-programmed network add +30.9 to +37.7 points of noise robustness while contrast moves by $-$0.28 to +0.03 (selection split, \cref{fig:stack}a).

The earlier capability is kept in the computation itself. None of the six contrast programs writes the two units of the noise gate or blocks 4 and 6 (\cref{fig:stack}b): constructed from the edited network, the earlier program and checks on the earlier condition, they write around the installed gate.

A counterfactual rerun shows why the current state of the network matters.
A contrast program written for the original network, applied after the noise program, frees the two block-2 units that hold the noise gate, because they are idle on clean images, and breaks 1,224 of 4,000 noisy check images; freeing those units accounts for 81\% of its noise loss (selection split, \cref{fig:stack}a,b).

%% file: fig/fig_sessions.tex
\begin{figure*}[t]
\centering
\includegraphics[width=\linewidth]{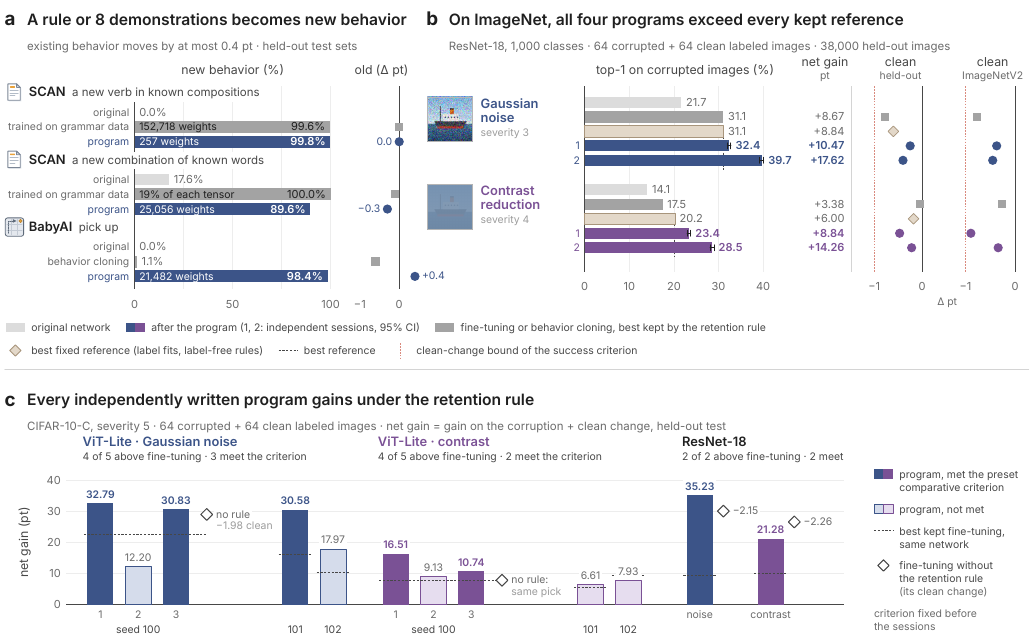}
\caption{\textbf{Weight programs install new capabilities and keep existing behavior.}
(a)~SCAN (a new verb in known compositions; a new combination of known words) and BabyAI before and after the program, with the weights each changes; gray, a rule-to-data reference trained on commands generated from the grammar (SCAN) and the best behavior cloning kept by the retention rule (BabyAI).
(b)~ImageNet ResNet-18: top-1 under corruption (95\% intervals), net gain, and the change in clean top-1 on held-out ImageNet and on ImageNetV2; dashed red, the clean bound of the success criterion.
(c)~Net gain of every CIFAR-10 program, 10 of 12 above the best kept fine-tuning on its network; light bars, programs that do not meet the prespecified comparative criterion; diamonds, fine-tuning without the retention rule.
Held-out test sets.}
\label{fig:sessions}
\end{figure*}

%% file: fig/fig_mech.tex
\begin{figure*}[t]
\centering
\includegraphics[width=\linewidth]{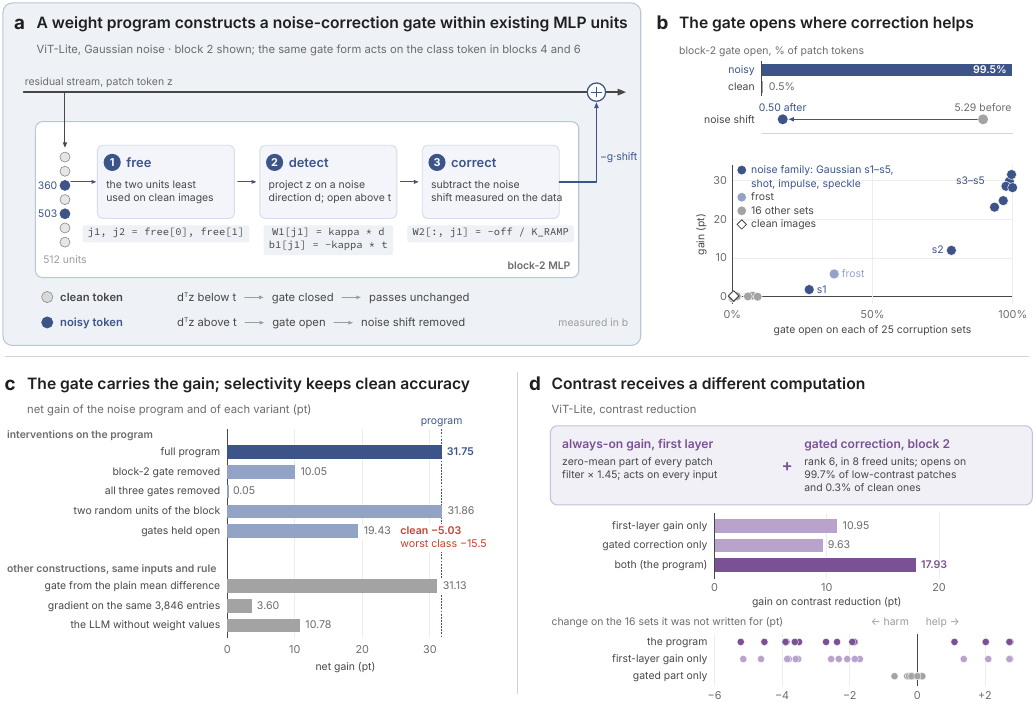}
\caption{\textbf{Weight programs install identifiable, causally testable computations.}
(a)~The noise program constructs a gate within existing MLP units that removes the noise shift (schematic).
(b)~The gate opens on noisy, not clean, inputs, and its activation tracks the corruptions where the program helps.
(c)~Causal interventions and alternative constructions identify the gated computation, more than the choice of units, as the source of this program's gain (\cref{tab:controls}).
(d)~The contrast program combines an input-wide gain adjustment with a selective correction; the two components have distinct target-condition and off-target effects (\cref{fig:range}).
Selection split.}
\label{fig:mech}
\end{figure*}

%% file: fig/fig_reuse.tex
\begin{figure*}[t]
\centering
\includegraphics[width=\linewidth]{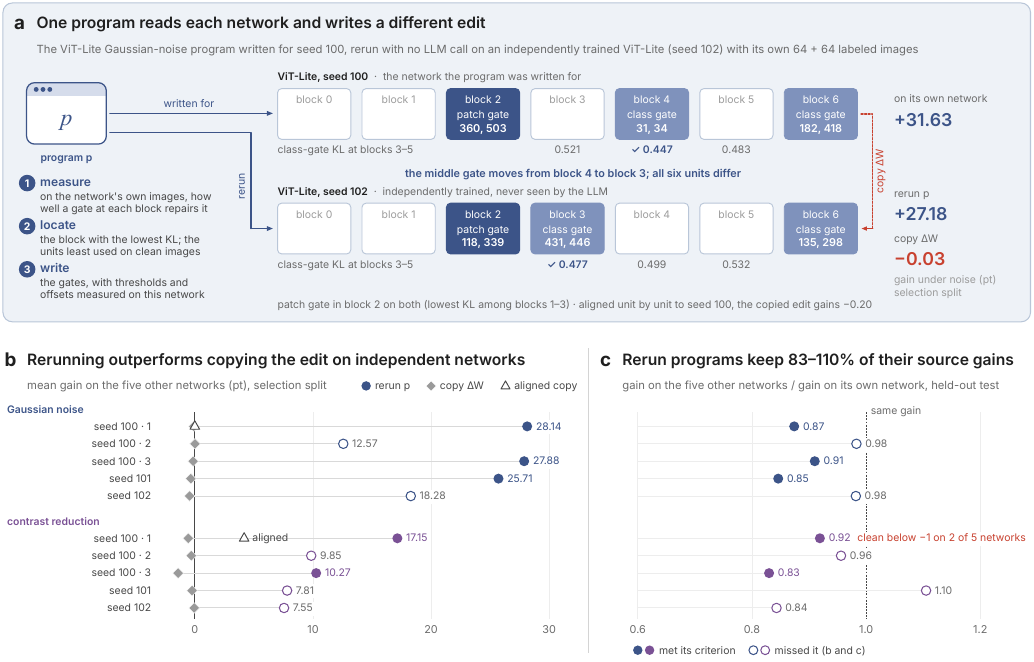}
\caption{\textbf{Executable weight programs reconstruct effective edits across independently trained networks.}
(a)~Rerun on an independently trained ViT-Lite, the noise program places its gates anew and keeps its gain, while its fixed weight edit, copied unchanged, produces no gain on this network (recorded placements, schematic).
(b)~Ten programs, rerun or copied to five other networks (selection split).
(c)~Share of each program's own gain kept on the other networks (held-out test sets). For networks that share an initialization, see \cref{fig:reuse_r18}.}
\label{fig:reuse}
\end{figure*}

%% file: fig/fig_stack.tex
\begin{figure*}[t]
\centering
\includegraphics[width=\linewidth]{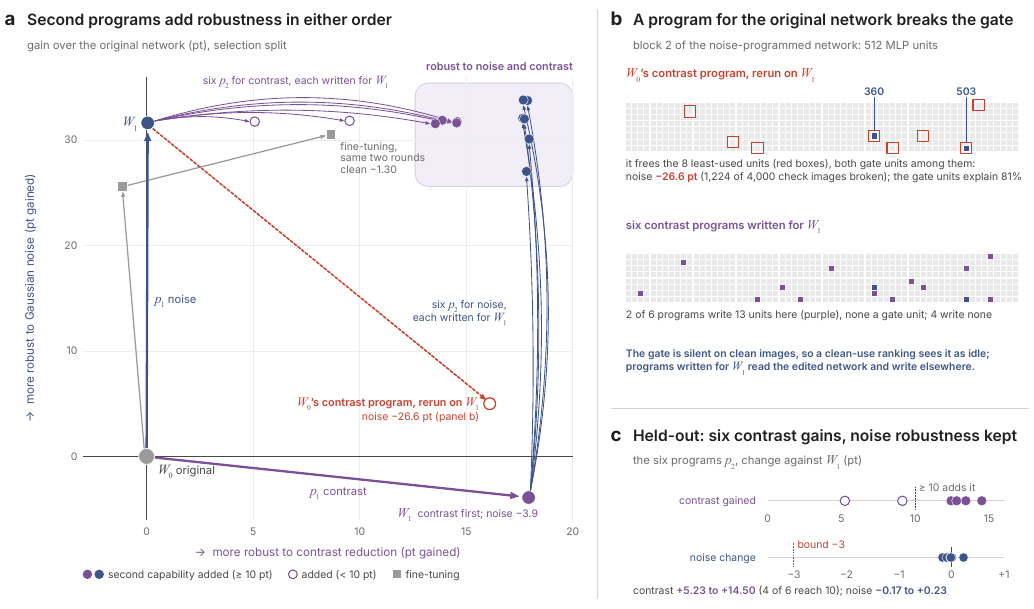}
\caption{\textbf{Programs written for an edited network add a capability and keep the earlier one.}
(a)~Gains over the original network in both orders: six alternative second programs per order, each written for $W_1$; red, the $W_0$ program rerun on $W_1$; gray, fine-tuning through the same two rounds (selection split).
(b)~Why a program written for the original network disrupts the installed noise gate when rerun on $W_1$: the block-2 units each program writes.
(c)~Held-out test sets for the noise-first order: all six programs add contrast robustness, four by at least the 10-point threshold fixed in advance.}
\label{fig:stack}
\end{figure*}

%% file: sec/6_discussion.tex
\section{Discussion}
\label{sec:discussion}

The central finding is that a trained neural network can be extended by an executable construction procedure whose output becomes part of the network's own computation.
The updated network performs the new capability on its own, while the program retains the procedure that built it from existing computation.
Capability installation therefore has an explicit implementation that can be tested by intervention and recomputed by rerunning the program.
This separates the reusable specification of a capability from the network-specific weight values that implement it.

These results reveal a mechanism of capability acquisition in which task knowledge is translated into new computation by recruiting and reorganizing operations already represented within a trained network.
Rules, examples and checks specify the desired behavior, while the network's weights and activations inform how it can be implemented: programs reconnect existing representations, recruit learned operations or add corrections for particular inputs.
In the noise-gating programs, preservation is built into the computation itself, in addition to the retention rule that selects them: the correction activates mainly on affected inputs and leaves clean representations nearly unchanged, and causal interventions separate the two roles, with the installed computation producing the robustness gain and its selective activation protecting clean accuracy.
The source of a new capability and the reason the measured existing behavior is kept can thus be examined directly and used to inform the design of later programs.

Reusability changes what can be retained after an update: not only the modified network, but also an executable description of how the modification was constructed.
Rerun on another network, that description recomputes an edit for the network's own weights; written for an edited network, a new program can account for what earlier programs installed.
As capabilities accumulate, previously installed computations become part of the foundation that later updates can use and preserve, and these properties motivate reusable program libraries and longer sequences of capability updates, each retained together with its checks.

Natural next steps are programs that construct representations a network does not yet have (\cref{sec:supp_bound}) and programs that extend foundation models, including LLMs, with new procedures or behaviors after deployment.
More broadly, executable weight programs provide a foundation for self-directed capability extension, in which proposed changes to a network's computation are evaluated independently, retained and built upon through further programs, each remaining code that can be read and tested.

Training establishes the initial computation of a neural network.
Executable weight programs offer a complementary route: task knowledge becomes part of the network's own computation while its measured existing behavior is kept, and each update is retained as a procedure that can be inspected, rerun and extended.
Weight programming thereby opens a path to cumulative capability acquisition, in which new computations are installed, reused and progressively extended within trained neural networks.

%% file: sec/7_supplement.tex
\section{Supplementary material}
\label{sec:supp}


This appendix provides implementation details, complete session-level results, additional controls, cross-condition analyses and exploratory experiments beyond the main protocol, in the order of \cref{sec:results}. \Cref{tab:main_dev,tab:examples_dev,tab:reuse_dev,tab:stack_dev} repeat \cref{tab:main,tab:examples,tab:reuse,tab:stack} with the numbers of the selection split.

\subsection{Protocol and implementation}
\label{sec:supp_impl}

The protocol specifies the task knowledge, the evaluation feedback and the retention bound for each setting (\cref{tab:protocol}).

\paragraph{What the protocol admits.}
Under the main protocol, weight updates are constructed from forward evaluations, network statistics and closed-form computations. Gradient-based and finite-difference optimization of objectives computed from the network's outputs (by automatic differentiation, hand-written backpropagation or finite differences), solvers on label-derived targets and iterated label-driven updates are excluded.
Three practices fall within the protocol: one overall edit strength found by a search against a label-free bound on the change of clean outputs, a one-shot closed-form bias set from the network's own logit margins (on BabyAI, where the only label-derived quantity is the index of the pickup action the task names), and a head bias set so that the edited network's mean logits on label-free inputs equal the original network's.
Manual code inspection identified three exploratory sessions that size their edits from finite differences of label-free output quantities: the ViT-Lite session given only the noise formula, and rounds 1 and 3 of a ResNet-20 stream. They fall outside the main protocol, are reported in \cref{sec:supp_fd}, and are not used to support the claims of \cref{sec:results}.

\begin{table*}[t]
\centering
\footnotesize
\setlength{\tabcolsep}{3pt}
\begin{tabular}{@{}>{\raggedright\arraybackslash}p{0.15\linewidth}>{\raggedright\arraybackslash}p{0.245\linewidth}>{\raggedright\arraybackslash}p{0.175\linewidth}>{\raggedright\arraybackslash}p{0.205\linewidth}>{\raggedright\arraybackslash}p{0.165\linewidth}@{}}
\toprule
Network and new condition & Task knowledge $I$ given to the LLM & Checks and retention rule & References & Selection and test sets \\
\midrule
ViT-Lite and ResNet-18; Gaussian noise or contrast reduction (CIFAR-10-C, severity 5) & 64 labeled images of the condition (16 the network misclassifies, 48 at random) and 64 labeled clean images, with the network's predictions; on the ViT-Lite also the lower levels of \cref{sec:supp_ladder} & $S_\text{new}$ and $S_\text{old}$ on 4,000 check images; $S_\text{old} \ge -20$ & eight fine-tuning methods, 8 configurations each; eleven fixed references (label fits, Tent, BN adaptation, SAR) & 4,000 images per condition; official CIFAR-10 and CIFAR-10-C test files \\
\addlinespace
Successive updates: ResNet-20 with Gaussian noise, defocus, snow and contrast in turn; ViT-Lite with Gaussian noise and contrast in either order & per round, 64 labeled images of the new condition and the code of the earlier programs; in one ViT-Lite arm also the list of entries round 1 changed & checks report every condition so far; stream clean change $\ge -60$ of 4,000, each earlier condition $\ge -20$ per round & ResNet-20: four fine-tuning streams re-selecting among 8 configurations each round, and fixed reference streams; ViT-Lite: criteria fixed in advance (\cref{sec:supp_stack}) & as above \\
\addlinespace
SCAN Transformer; add primitive (\emph{jump}) or around right & the interpretation grammar, 64 old commands with answers and the network's predictions, and on add primitive the training row of \emph{jump} & $S_\text{new}$ on a third of the official test commands, $S_\text{old}$ on 1,000 old commands; $S_\text{old} \ge -5$ & fine-tuning on the given rows; fine-tuning on commands generated from the grammar & a second third and 1,000 old commands; the last third and 1,000 old commands \\
\addlinespace
BabyAI policy; pick up & 8 demonstrations of the new mission (observation, instruction and expert action at each step) and 64 old steps, with the policy's action logits & rollouts of 500 new-mission and 500 old-mission episodes; $S_\text{old} \ge -3$ & the eight fine-tuning methods and behavior cloning of parts of the policy; full behavior cloning from 16 to 256 demonstrations & 1,000 episodes per mission; 1,000 further episodes per mission \\
\addlinespace
ResNet-18 on ImageNet (1,000 classes); Gaussian noise (severity 3) or contrast reduction (severity 4) of ImageNet-C & as for CIFAR-10, from held-out validation images & as for CIFAR-10, on 4,000 validation images each & eight fine-tuning methods; eleven fixed references (label fits, Tent, BN adaptation) & 4,000 validation images per condition; the other 38,000 validation images \\
\bottomrule
\end{tabular}
\caption{\textbf{What each task gives the LLM and the references.} A session is one LLM session with 8 checks per round and the retention rule of \cref{sec:method_install} with the bound shown, and the network is left unchanged when no candidate meets it. $S_\text{new}$ and $S_\text{old}$ count inputs fixed minus inputs broken, on the new condition and on old inputs. Every reference receives the same examples or rule and the same checks and is kept by the same rule. Matched checks are the external evaluation checks; label-scored work inside a session is counted as described in \cref{sec:supp_impl}. Analyses are read on the selection set, and each kept network is read once on the test set.}
\label{tab:protocol}
\end{table*}

\paragraph{Sessions and sandbox.}
The following measurements describe the cost of constructing a program, distinct from the cost of rerunning a kept program. The LLM works in a Python kernel of at most 150 cells, within 90 to 120 minutes (140 on ImageNet and 190 on the CIFAR-10 ResNet-18, whose checks take longer). A session takes 30 to 130 minutes, and the CIFAR-10, SCAN and BabyAI sessions cost US\$5 to 13 at API prices. Rerunning a kept program on a further network takes a few minutes with no LLM call.
A program runs in a sandbox with NumPy only, no network access and a 300\,s limit.
It may change at most 20\% of the weights, and the change to each tensor of rank two or more is at most half of that tensor's Frobenius norm. The per-tensor limit applies to the image classifiers but not to the SCAN Transformers or the BabyAI policy, whose edits write whole embedding or output rows.

\paragraph{Labels and computation in a session.}
The 8 external checks count labeled check images at every level of information, including the levels with condition labels withheld and with the name only.
A single-round session scores at most 8 candidate weight edits against the labeled examples in $I$, as each fine-tuning method tries 8 configurations.
Two kinds of label-scored work are not counted toward this cap: predictors that change no weight of the network, such as output offsets or nearest-class-mean probes on its features, and selection a program makes inside its own run.
This counting rule was set on 2026-11-06, after two sessions had been read: the noise-formula session (\cref{sec:supp_fd}), which scores 5 candidate edits under the rule and 9 settings with its four logit offsets, and the CIFAR-10 ResNet-18 contrast session, which scores 6 under the rule and 9 with its three nearest-class-mean probes. The rule applies to every session, and the automatic audit counts both kinds, so its counts are recorded beside the hand counts.
The cap applies to single-round sessions. A session that writes a later round scores 0 to 44 settings against that round's labeled examples, including threshold sweeps, and the fine-tuning streams fit the same labels.
On SCAN, whose grammar gives the answer to any command, the LLM may test candidates against answers it derives from the rule.
A kept program computes from the network it runs on each time it runs, with forward passes, activation statistics, SVD and PCA on the inputs in $I$. The contrast program written without weight values (\cref{sec:mechanism}) also selects among eight configurations by cross-validation on the labeled examples each time it runs.
The audit flags gradients, solvers on label-derived targets, iterated label-driven updates and the number of settings scored against labels.

\paragraph{Data and networks.}
Corruptions are generated with the official CIFAR-10-C code on held-out CIFAR-10 training images. On the clean test images this code reproduces the official files exactly for defocus and contrast and within 1.2\% of their median pixel deviation for noise and snow.
The ViT-Lite follows the authors' CIFAR-10 recipe. Gaussian noise leaves it 35.55\% accuracy and contrast reduction 58.88\%, while defocus and snow cost it 12 and 13 points.
The ResNet-18 receives a new 10-way head and is fine-tuned in full with inputs upsampled to $128 \times 128$.
The BabyAI policy is a feedforward network with FiLM conditioning on the instruction, trained by imitation on 50,000 expert demonstrations of the go-to mission in one room.
Each SCAN Transformer (width 128, 4 layers) is trained on the official training commands of its split except 3,000 held out for the old checks, selection and test commands.
The reuse networks are trained from seeds 101 to 105, and from seed 100 for programs written on seeds 101 and 102.

\subsection{Comparison with fine-tuning}
\label{sec:supp_ft}

\input{tables/tab_behaviour}
\input{tables/tab_main}

This section evaluates how the relative performance of weight programs and fine-tuning changes with the data budget, the information about the corruption and the permitted loss of clean accuracy.
On the ViT-Lite, no fine-tuning variant reaches the program's net gain within the retention rule; among the evaluated variants, matching or exceeding it requires the true corruption model and no clean bound, or 64 times the labeled images or more (\cref{tab:examples}, \cref{fig:candidates}, \cref{fig:f2}). On BabyAI, no configuration of the eight fine-tuning methods keeps an edit within the retention rule, and full behavior cloning keeps no edit even from 256 demonstrations (selection split).

\paragraph{Same examples and checks.}
Under the matched-data and retention protocol, the ViT-Lite programs reach net gains of +32.79 and +16.51, against +22.78 and +7.89 for the best of the eight evaluated methods. Tuning the recipe with warmup, augmentation and minibatches keeps +21.34 and +7.58.

\paragraph{A corruption model.}
Images synthesized from a corruption model estimated from the 64 images add no independently labeled observations, although they carry the assumptions of the fitted model. Fine-tuning with them keeps no edit under Gaussian noise and reaches +10.00 under contrast reduction.
The true corruption model adds the process that generates the condition, which the programs are not given. With it, fine-tuning keeps no edit under Gaussian noise and reaches +10.23 under contrast reduction.

\paragraph{Trade-offs under relaxed bounds.}
Relaxing the clean-accuracy bound changes the attainable trade-off between gain and retention: weight programs hold their largest advantage under tight bounds, and the evaluated fine-tuning configurations reach larger gains when larger clean-accuracy changes are permitted. Without the rule, its best configurations reach +29.11, +30.18 and +26.70 on the test sets at clean changes of $-$1.98 to $-$2.26, the last above the CIFAR-10 ResNet-18 contrast program (\cref{tab:main}); when the clean-loss bound allows 2 points, fine-tuning passes the program by 0.78 points under Gaussian noise on the ViT-Lite and by 4.95 under contrast reduction on the CIFAR-10 ResNet-18, at clean changes of $-$1.58 and $-$2.38 (selection split). On the ViT-Lite under Gaussian noise, the tuned recipe reaches +31.13 at $-$2.23 and the true corruption model +32.97 at $-$2.78. Under contrast reduction the best variant reaches +11.49. \Cref{fig:candidates}b--e shows every checked configuration on the selection split.

\paragraph{More labeled images.}
On the ViT-Lite, with $N$ labeled images of the condition and as many clean ones, the evaluated full fine-tuning recipe first reaches the program's net gain at $N = 4{,}096$ under Gaussian noise, 64 times the program's 64, and between $N = 4{,}096$ and $40{,}000$ under contrast reduction (\cref{fig:f2}, selection split). Under noise the retention rule sets this amount: at $N = 64$ to $1{,}024$ the same fine-tunes gain +30.7 to +38.9 but lose more clean check images than the rule allows.

\input{tables/tab_examples}
\input{tables/supp_main_dev}
\input{tables/supp_examples_dev}
\input{fig/fig_f2}
\input{fig/fig_candidates}

\subsection{Every session}
\label{sec:supp_sessions}

\input{tables/tab_sessions}

\Cref{tab:sessions} reports every independent ViT-Lite session, so that installation, the comparison with fine-tuning and reuse can be assessed separately. Of the ten programs, eight exceed the best kept fine-tuning on their own network, five meet the prespecified success criterion, and all ten keep 0.83 to 1.10 of their gain when rerun on the other five networks.
Under Gaussian noise three of five sessions meet the criterion, and under contrast reduction two of five. On the seed-100 network every contrast program exceeds the best kept fine-tuning (+9.13 to +16.51 against +7.89), and two of three noise programs do (+30.83 and +32.79 against +22.78).
For the first Gaussian-noise session on the seed-100 network, an expanded reference grid was fixed after the session began but before its first check; the program meets both the original and the expanded criterion.

Across these sessions, higher net gains are associated with the computation the program writes.
The three Gaussian-noise programs that meet the criterion build a saturating gate from MLP units with little effect on clean outputs; the other two apply Wiener shrinkage without an MLP gate.
Every contrast program raises the first-layer filters. The two that meet the criterion also turn little-used capacity into a correction, gated in freed MLP units or written into a block-0 attention head that subtracts the image's mean color. The three that add no such correction gain +6.77 to +9.51.
Rerun on the other five networks, every program keeps clean accuracy within 1 point except the first contrast program, which loses more on two of five. Every second-round session is in \cref{tab:rounds}.
Each session used 5 to 8 checks and cost US\$7.12 to 13.08.

\subsection{ImageNet}
\label{sec:supp_imagenet}

On torchvision's ResNet-18 with its 1,000 ImageNet classes, all four programs meet the prespecified criterion on the held-out test set and exceed every fine-tuning method; one contrast session exceeds the clean-change bound on the selection split (\cref{tab:imagenet}).

\paragraph{Setup.}
The network is torchvision's IMAGENET1K\_V1 ResNet-18, unchanged. The conditions are ImageNet-C's Gaussian noise and contrast functions, run as published on the $224 \times 224$ crops. The evaluated severities were selected with the original network on a separate split of 1,000 validation images that the test set excludes: severity 3 for Gaussian noise and 4 for contrast reduction, the highest severities that cost at least 15 points and keep at least 17\% of the accuracy above chance (at severity 5 the network keeps 1.4\% and 3.6\%).
Each session follows the protocol of the CIFAR-10 sessions, with the same 64 labeled images of each kind, 8 checks on 4,000 condition and 4,000 clean check images and the bound of 20 clean check images. The references are the eight fine-tuning methods and eleven fixed references of the CIFAR-10 ResNet-18. Their grids were extended once toward smaller steps, by a rule fixed before any session, since larger steps cost this network more clean accuracy than the bound allows.

\paragraph{Results.}
Under Gaussian noise the two programs reach net gains of +10.47 and +17.62 on the test set, against +8.67 for the best fine-tuning, and under contrast reduction +8.84 and +14.26, against +3.38.
One contrast session has a clean change of $-$1.13 points on the selection split, on which the kept programs were read before the test set, and $-$0.47 on the test set; the other three programs meet the criterion on both splits.
Full fine-tuning on 31 times the labeled images, 2,000 of each kind, keeps no configuration within the bound; its best configurations reach +20.53 and +30.60 at clean changes of $-$1.10 and $-$1.95 (selection split).

\paragraph{Neighboring severities and ImageNetV2.}
On 20,000 test images each program gains most at its own severity and keeps a positive gain at the neighboring ones. The largest change on the other condition is $-$3.36 points, for the first contrast program under Gaussian noise.
On ImageNetV2~\cite{recht2019imagenetv2}, 10,000 new images collected like the validation set, the programs change clean accuracy by $-$0.34 to $-$0.89 points and the kept fine-tuning by $-$0.77 and $-$0.27.

\paragraph{What the programs write.}
The Gaussian-noise programs estimate the noise from pixel statistics of the given images, synthesize noisy copies of the clean ones and remove the noisy-minus-clean shift of activations in the deep half of the network: the first at the inputs of two residual stages and of the classifier, folded into their weights and BatchNorm statistics, and the second as a rank-one term in eight convolutions that a whitened noise detector switches on.

The contrast programs measure the contrast factor from pixel statistics and turn the two least-used units of several residual blocks into gates that open on low-contrast inputs and add the activation shift that restores them. The first gate opens from the 75th percentile of the clean projections, so a quarter of clean locations receive part of the correction, and the second only above the 99.5th, which separates their clean costs.

\input{tables/tab_imagenet}

\subsection{Less information}
\label{sec:supp_ladder}

On the ViT-Lite under Gaussian noise, a program meets the bar fixed before these sessions from labeled examples and with the labels of the condition images withheld (\cref{fig:ladder}).
Each information condition is evaluated in one session on the seed-100 network, read on the selection split against the bar of +17.075. With labeled examples the program reaches +31.75, and with the condition labels withheld +33.37. From the capability's name alone, a program declared in a session without checks and scored once afterwards reaches +15.10, and a session with the same sentence and 8 checks +9.43. The session given only the noise formula is in \cref{sec:supp_fd}. Together, these sessions show that programs can be constructed from several forms of task information, including condition images without their labels and a one-sentence description of the capability. Test-time adaptation on the same unlabeled images, kept by the same rule, leaves the network unchanged (Tent~\cite{wang2020tent}) or reaches +0.025 (SAR~\cite{niu2023sar}).
On SCAN, from the grammar, the programs reach net gains of +99.84 and +72.02 exact-match points, where fine-tuning on the same rows keeps +1.21 and +2.61 (selection split). Fine-tuning on commands generated from the grammar reaches +99.71 and +83.68 by changing 152,718 entries. Because the grammar can generate commands in the evaluation set, grammar-generated fine-tuning is a rule-to-data reference rather than a standard held-out supervised baseline. Both methods are given the same rule. The programs write the rule into the network directly, with 257 and 25,056 weights.

\input{fig/fig_ladder}

\subsection{What programs write and what carries the gain}
\label{sec:supp_what}

\Cref{tab:what} characterizes the computation each reported program implements, from its code and its stated hypothesis; for four programs, interventions on the network test which parts carry the gain.
In the Gaussian-noise program, removing the block-2 gate lowers the net gain from +31.75 to +10.05, removing all three gates returns the network to its original accuracy, and holding them open costs 5.03 points of clean accuracy (\cref{fig:mech}c).
In the contrast program, the first-layer gain alone adds +10.95 and the gated correction alone +9.63, and together they reach +17.93 (\cref{fig:mech}d).
The SCAN add-primitive program changes the embedding row of \emph{jump} (128 entries), the output row of its action (128) and that row's bias (1), and leaves every layer between them unchanged. The embedding change alone installs nothing, the output change alone reaches 26.90\% exact match, and together they reach 99.84\% (selection split). With the embedding change alone, every action slot of \emph{jump} is decoded as \emph{walk}, \emph{look} or \emph{run}, and the hidden states there lie at the center of the states the original network computes for those three words, at a fifth of their spread at the input of the output layer. The new behavior thus depends on coordinated input and output edits, neither of which alone reproduces the compositional capability: the new primitive enters through the composition the network already computes, and the output row names it.
In the BabyAI program, a rule written outside the network from its three parts matches the edited policy on each of 2,000 episodes (\cref{sec:mechanism}).

\Cref{tab:controls} gives every control of the two ViT-Lite programs discussed in \cref{sec:mechanism}, and one more alternative: an unconditional feature alignment, which applies its correction to every input, reaches +7.18 under noise, against +31.75 for the gated program.

\begin{table*}[t]
\centering
\small
\setlength{\tabcolsep}{4pt}
\begin{tabular}{@{}>{\raggedright\arraybackslash}p{0.22\linewidth}>{\raggedright\arraybackslash}p{0.64\linewidth}r@{}}
\toprule
Network, condition, information & What the program writes & Weights \\
\midrule
ViT-Lite, Gaussian noise, labeled\textsuperscript{*} & frees two little-used MLP units in each of blocks 2, 4 and 6 and rewrites them as ramp gates that subtract the measured noise offset from tokens above the clean range & 3,846 \\
ViT-Lite, Gaussian noise, condition labels withheld & estimates the noise strength from the images' excess neighbor-difference energy and chains saturating gates in the least-used MLP units of blocks 2 to 4 and the last block, each triggered by a flag the previous gate writes & 19,724 \\
ViT-Lite, Gaussian noise, name only & gives the 16 block-0 MLP units most driven by noise twin units that cancel the mean the noise induces through the GELU & 24,848 \\
ViT-Lite, contrast reduction, labeled\textsuperscript{*} & raises the zero-mean part of the patch-embedding filters by a factor of 1.45 and adds a gated rank-6 correction in eight freed units of block 2 & 16,648 \\
ResNet-18 (CIFAR-10), Gaussian noise, labeled & estimates the noise level from the images' high-frequency energy, applies a Wiener shrinkage to each convolution in forward order and shifts the following BatchNorm means to keep clean pre-activations; no gate & 1,718,057 \\
ResNet-18 (CIFAR-10), contrast reduction, labeled & raises the zero-mean part of the stem convolution by 35\% and adds gates on a whitened low-contrast direction in two late residual blocks & 37,192 \\
ResNet-18 (ImageNet), Gaussian noise, labeled & estimates the noise from pixel statistics of the given images, synthesizes noisy copies of the clean ones and removes the noisy-minus-clean activation shift at the inputs of two residual stages and of the classifier, folded into their weights and BatchNorm statistics & 2,152,928 \\
ResNet-18 (ImageNet), contrast reduction, labeled & measures the contrast factor from pixel statistics and turns the two least-used units of several residual blocks into gates that open on low-contrast inputs and add the activation shift that restores them & 33,176 \\
BabyAI policy, pick up, 8 demonstrations\textsuperscript{*} & copies the embeddings of \emph{go} and \emph{to} into \emph{pick} and \emph{up}, builds from dead units an exact detector for a faced cell of the named color and type that triggers the pickup action, and adds a second detector that stops the policy from pressing forward into a blocking object & 21,482 \\
SCAN Transformer, add primitive, rule\textsuperscript{*} & moves the embedding of \emph{jump} to the centroid of those of \emph{walk}, \emph{look} and \emph{run} and adds an output row and bias for its action that win where the actions of those three tie & 257 \\
SCAN Transformer, around right, rule & writes rank-one edits of attention values in blocks 2 to 4 along the network's own right-minus-left turn direction & 25,056 \\
\bottomrule
\end{tabular}
\caption{\textbf{What each kept program writes.} The program reported in \cref{sec:results} for each network, condition and level of information, described from its code and its stated hypothesis, with the number of weights it changes. \textsuperscript{*}Programs whose parts were tested by intervention on the network (\cref{sec:supp_what}); the other descriptions come from code and stated hypotheses.}
\label{tab:what}
\end{table*}

\input{tables/tab_controls}

\subsection{Where the programs act}
\label{sec:supp_range}

This analysis examines whether an installed computation affects only its target condition or also changes behavior under other corruptions. We read the four kept programs of the ViT-Lite and the CIFAR-10 ResNet-18, and the kept fine-tuning of the same conditions, on clean images and 25 corrupted versions of the test set: each CIFAR-10-C corruption at severity 5, and Gaussian noise and contrast at severities 1 to 4 (\cref{fig:range}). The effect directions are stable across splits: 197 of the 208 changes carry the same label on the test sets as on the selection split, and none moves between help and harm.
Each Gaussian-noise program helps on the whole noise family, at every severity and under shot, impulse and speckle noise, and on frost. The ViT-Lite program changes no other set by more than 0.27 points and harms none, where its fine-tuning harms 6. Its block-2 gate opens on 27\% to 100\% of patch tokens on the sets the program helps and on at most 9\% elsewhere (Spearman 0.77 between gate share and change over the 25 sets, selection split). The ResNet-18 program, which writes no gate, helps on 14 sets and harms 5: contrast at severities 2 to 5, and fog.
In each contrast program, the always-on first-layer gain and the gated part differ in reach. Together they help on the program's own condition, the three blurs and fog, and harm every noise set: 12 sets on the ViT-Lite and 10 on the ResNet-18, against 2 and 1 for fine-tuning. Applied alone to the original network, the gain reproduces 0.92 to 1.25 of the program's change on each of these sets, and the gated part alone changes none of them by more than 0.68 points (selection split). Clean accuracy and robustness to other corruptions are thus separate properties, and the first-layer gain governs the second. Without the gain, each program keeps 0.54 (ViT-Lite) and 0.67 (ResNet-18) of its gain on its own condition.
On their own condition at severity 5, all four programs exceed their fine-tuning, by 9.2 to 25.7 points.

\input{fig/fig_range}

\subsection{Rerunning programs on other networks}
\label{sec:supp_reuse}

This section gives the per-network measurements behind \cref{sec:reuse} and examines the role of weight coordinates (\cref{tab:reuse}).
On the five ViT-Lites trained from other seeds, the Gaussian-noise program gains +27.55 to +30.54, 0.87 of its gain on its own network, and the contrast program keeps 0.92. Fine-tuning recipes retain comparable fractions of their source-network gains but reach lower absolute gains in the evaluated transfers: the best fine-tuning configuration of the seed-100 network, applied unchanged, keeps 0.89 and 0.98 of a lower gain (+20.79 and +8.09), and repeating the fine-tuning selection on each network gives +10.44 and +6.57.
Three measurements locate the transfer advantage. First, the program reselects locations and parameters on each network: on one network the noise program gates blocks 2, 3 and 6 instead of 2, 4 and 6, and across the five networks its block-2 threshold ranges from $-$7.0 to $-$11.3, against $-$10.35 on the original. Second, the tested activation-based alignment does not recover a fixed edit: after unit-by-unit alignment the units the program changes coincide with the original's in 3 of 28 positions across the five networks, and the aligned copies of the first program of each condition recover $-$0.01 and +4.17 (selection split). Third, a fixed edit transfers well when networks share coordinates. Added unchanged to independently trained ViT-Lites, the edits of the ten programs gain $-$1.42 to +0.01 (selection split, \cref{tab:sessions}). On five ResNet-18s fine-tuned from the same ImageNet weights, the copied edits carry 96\% and 87\% of the rerun gain, and rerunning gains more (\cref{fig:reuse_r18}, selection split).

\input{tables/tab_reuse}
\input{tables/supp_reuse_dev}
\input{fig/fig_reuse_r18}

\subsection{Successive updates}
\label{sec:supp_stack}

Across twelve second-round sessions on the ViT-Lite, the earlier condition is preserved within the reported tolerance, and the size of the added capability differs by update order and session (\cref{tab:rounds}).
On the ViT-Lite, the earlier condition moves by $-$0.28 to +0.23 points in every second-round session, and clean accuracy stays within 1.47 points of the original network.
With Gaussian noise first, four of six sessions add at least 10 points of contrast robustness. With contrast first, all six add at least 10 points of noise robustness.
Giving a session the list of entries round 1 changed leaves this rate unchanged (two of three and three of three, as without it).


\input{tables/tab_rounds}

\subsection{Capabilities beyond the network's identified computation}
\label{sec:supp_bound}

The programs of \cref{sec:results} build each capability on computation the network already has. Two additional settings probe cases in which the target capability is not directly supported by the computations or representations identified in the original network: a new operation that must also run inside the old ones, and new classes that the network's features do not separate. In both, the program adds the capability as a separate component: a lookup table beside the old operations, and prototype rows on the base features. Each is one session, read on the selection split.

\subsubsection{A new operation on a synthetic sequence task}
\label{sec:supp_seq}

From a symbolic rule, a weight program installs a new operation with 100\% exact match on inputs of length 4 to 8 and 92.75\% on inputs of length 9 to 12, while preserving the original operations.
The sequence Transformer (width 64, 205,008 weights) is trained on eight symbolic operations and receives a new operation bound to a symbol permutation. The LLM is given only its symbol table, with no row of the new operation, and 64 rows of the old ones.
It does so in one check, from 0\%; the old operations stay at 100\%, and none of their 4,000 selection rows changes.
Changing 4,465 weights (2.18\%), it flags the new operation along the residual stream's lowest-variance direction, routes the flag to the output positions through one attention head, and turns 16 little-used last-layer MLP units into a lookup table.
The installed operation does not run inside the existing operations: composed with an old operation, the written lookup reaches 5.95\% exact match, whereas fine-tuning on rows generated from the same rule reaches 100\% on both lengths and on the composition, at $-$0.22 points on the old operations.
On SCAN, where the program writes into the input and output rows and leaves the composition to the network, the new primitive runs inside the old compositions (\cref{sec:supp_what}).

\subsubsection{New classes on a small CNN}
\label{sec:supp_fscil}

This experiment evaluates whether a weight program can add five new classes to a classifier whose features do not clearly separate them: four of the five new classes lie inside old classes in the network's features.
On the few-shot class-incremental protocol of CEC~\cite{zhang2021cec} on CIFAR-100 (60 base classes, then five new classes per session from five labeled images each), one session wrote the five classes of session 2 into a ResNet-20 with a cosine head.
The kept program writes prototype rows with margins and keeps the base classes within 0.78 points. It reaches 17.20\% on the new classes and a class-balanced 64.55\% on all seen classes, against 21.70\% and 65.11\% for imprinted prototypes, the strongest reference, and a criterion of 26.70\% and 65.11\%.
In this single session, the generated programs stay within a prototype-based update family and do not reach the criterion fixed before the session.

\subsection{Sessions that size edits from finite differences}
\label{sec:supp_fd}

Three exploratory sessions set the size of their edits from finite differences of a label-free output quantity, which the main protocol excludes: the ViT-Lite session given only the noise formula, and rounds 1 and 3 of a ResNet-20 stream. Round 4 of the stream starts from the values round 3 set. The automatic audit, which checks finite differences of label-scored losses, did not detect these operations; they were identified by subsequent code inspection. These exploratory results are excluded from the evidence supporting the protocol-constrained claims of \cref{sec:results}.

\paragraph{The noise formula.}
In an exploratory session given the formula that generates Gaussian noise and 64 labeled clean images, with no noisy image, the program synthesizes noisy copies of the clean images and shrinks the low signal-to-noise components of the ViT-Lite's patch embedding and the MLP units that respond to noise. It sets each shrink fraction from a finite-difference quotient of the divergence between the network's outputs on clean and noisy copies, scaled by one overall factor found against the clean damage.
Changing 34,304 weights, it reaches a net gain of +18.27 at a clean change of $-$0.46 on the test set (+18.15 on the selection split). Fine-tuning on images generated from the same formula keeps no edit within the retention rule and reaches +32.07 at $-$2.83 without it (selection split).

\paragraph{The ResNet-20 stream.}
ResNet-20 receives Gaussian noise, defocus, snow and contrast in turn, with 64 labeled examples of each, one new program per round written for the network the earlier rounds produced, and a clean budget of 60 of 4,000 check images for the whole stream.
Round 1's program sets each unit's shrink fraction from how much ablating the unit reduces a label-free output divergence on noisy inputs, divided by the square of its clean cost. Round 3's session adapts each BatchNorm layer alone, measures the change of label-free logit projections, and solves each layer's strength from it; the kept program carries the eleven values.
The round-1 program also reads the wall clock and stops its sweep over units after 220\,s. The reported stream uses the recorded round-1 execution, which examined 524 of its 688 units under machine load; a full sweep produces a different $W_1$. Rounds 2 to 4 and the selection among separately written programs below start from this recorded network state, so the stream's later rounds are tied to that execution.

In this exploratory stream, which includes finite-difference operations excluded from the main protocol, the mean gain over the four corruptions after four rounds is +10.61 at a clean change of $-$1.66, a stream score of +8.95 against +5.15 for the best of four fine-tuning streams under the same feedback, and every corruption ends at or above its gain after its own round (\cref{tab:stack}, \cref{fig:stream}).
The round-2 program, written for defocus, improves all four corruptions (defocus +10.25, Gaussian noise +9.46, contrast +6.86 and snow +1.90 points over round 1).
Later rounds add smaller gains once the stream has consumed its clean-accuracy budget: rounds 1 and 2 spend 25 and 35 of the 60 images, which leaves rounds 3 and 4 a floor of 0 and $-$1 images, and these rounds add +0.78 and +0.53 points on their own corruptions (selection split).
Selecting among the four programs written separately for the original network reaches a stream score of +3.48, and the fixed references under the same feedback +4.84 (selection split).

\input{tables/tab_stack}
\input{tables/supp_stack_dev}
\input{fig/fig_stream}

%% file: tables/tab_behaviour.tex
\begin{table*}[t]
\centering
\small
\setlength{\tabcolsep}{4pt}
\begin{tabular}{@{}l>{\raggedright\arraybackslash}p{0.19\linewidth}cccc>{\raggedright\arraybackslash}p{0.17\linewidth}cc@{}}
\toprule
& & \multicolumn{4}{c}{Weight program (\method)} & \multicolumn{3}{c}{Other route from the same information} \\
\cmidrule(lr){3-6}\cmidrule(l){7-9}
Setting & Given & new, base & new, after & old & weights & route & new & old \\
\midrule
SCAN, add primitive & grammar, 64 old commands, the \emph{jump} row & 0.0\% & 99.8\% & 0.0 & 257 & grammar-generated fine-tuning & 99.6\% & 0.0 \\
SCAN, around right & grammar, 64 old commands & 17.6\% & 89.6\% & $-$0.3 & 25,056 & grammar-generated fine-tuning & 100.0\% & $-$0.1 \\
\midrule
BabyAI, pick up & 8 demonstrations, 64 old steps & 0.0\% & 98.4\% & +0.4 & 21,482 & behavior cloning, embeddings and head & 1.1\% & $-$0.6 \\
\bottomrule
\end{tabular}
\caption{\textbf{Programs enable behaviors the original networks did not perform.} New: exact match on the SCAN commands that contain the new primitive or the new combination, or success on the new BabyAI mission, before and after the update. Old: change in points on the old commands or the old mission. Weights: entries the program changes. Grammar-generated fine-tuning trains on commands generated from the grammar and mixed with the old ones, and changes 152,718 entries on add primitive and a fixed 19\% of each tensor on around right; its generated commands include some test commands, so it is a rule-to-data reference rather than a held-out supervised baseline. Behavior cloning of the embeddings and the head is the strongest reference kept by the program's rule on BabyAI. Fine-tuning on the given rows alone is in \cref{sec:examples}. Held-out test sets.}
\label{tab:behaviour}
\end{table*}

%% file: tables/tab_main.tex
\begin{table*}[t]
\centering
\small
\setlength{\tabcolsep}{4.5pt}
\begin{tabular}{@{}llccccccr@{}}
\toprule
& & \multicolumn{3}{c}{Weight program (\method)} & \multicolumn{2}{c}{Fine-tuning, best of 8} & Fixed refs. & Weights \\
\cmidrule(lr){3-5}\cmidrule(lr){6-7}\cmidrule(lr){8-8}
Network & Condition & gain & clean & net gain & kept & no bound & kept & changed \\
\midrule
\multicolumn{9}{@{}l}{\emph{ImageNet, 1,000 classes, corruptions of ImageNet-C at severity 3 (noise) and 4 (contrast)}} \\
ResNet-18 & Gaussian noise & +10.72 & $-$0.25 & \textbf{+10.47} & +8.67 & -- & +8.84 & 2,152,928 \\
ResNet-18 & contrast reduction & +9.31 & $-$0.47 & \textbf{+8.84} & +3.38 & -- & +6.00 & 33,176 \\
\midrule
\multicolumn{9}{@{}l}{\emph{CIFAR-10, corruptions of CIFAR-10-C at severity 5}} \\
ViT-Lite & Gaussian noise & +32.68 & +0.11 & \textbf{+32.79} & +22.78 & +29.11 ($-$1.98) & +16.48 & 3,846 \\
ViT-Lite & contrast reduction & +17.45 & $-$0.94 & \textbf{+16.51} & +7.89 & +7.89 ($-$0.34) & +8.25 & 16,648 \\
ResNet-18 & Gaussian noise & +35.61 & $-$0.38 & \textbf{+35.23} & +9.47 & +30.18 ($-$2.15) & +9.22 & 1,718,057 \\
ResNet-18 & contrast reduction & +21.43 & $-$0.15 & \textbf{+21.28} & +10.01 & +26.70 ($-$2.26) & +13.51 & 37,192 \\
\bottomrule
\end{tabular}
\caption{\textbf{Under matched task data, checks and clean-accuracy retention, weight programs achieve larger net gains than the evaluated fine-tuning methods.} Accuracy changes in points on the held-out test sets. ImageNet: torchvision's ResNet-18 with its 1,000 classes (\cref{sec:supp_imagenet}). CIFAR-10: ViT-Lite-7/4 trained on CIFAR-10 and an ImageNet-pretrained ResNet-18 fine-tuned on CIFAR-10. Net gain is the gain on the new condition plus the clean change. Every method receives the same 64 labeled images of the condition and 64 labeled clean images (the fine-tuning variant without clean images fits only the 64 of the condition) and the same 8 checks. Kept: the checked configuration with the best check score among those losing at most 0.5\% of clean check images. No bound: the configuration with the best check score at any clean loss, with its clean change in brackets (not run on ImageNet). Fixed refs.: the best of eleven label fits and label-free rules, kept by the same rule. Weights: entries the program changes. Each program row is the first session on its network; every ImageNet session is in \cref{tab:imagenet} and every ViT-Lite session in \cref{tab:sessions}. Bold: the highest net gain among the methods kept by the rule.}
\label{tab:main}
\end{table*}

%% file: tables/tab_examples.tex
\begin{table*}[t]
\centering
\small
\setlength{\tabcolsep}{5pt}
\begin{tabular}{@{}llccccccc@{}}
\toprule
& & \multicolumn{3}{c}{Gaussian noise} & & \multicolumn{3}{c}{Contrast reduction} \\
\cmidrule{3-5}\cmidrule{7-9}
& & gain & clean & net gain & & gain & clean & net gain \\
\midrule
\multicolumn{2}{@{}l}{Weight program (\method)} & \textbf{+32.68} & \textbf{+0.11} & \textbf{+32.79} & & \textbf{+17.45} & $-$0.94 & \textbf{+16.51} \\
\midrule
\multirow{4}{*}{\shortstack[l]{Fine-tuning,\\kept by the\\same rule}}
& best of 8 methods & +23.41 & $-$0.63 & +22.78 & & +8.23 & $-$0.34 & +7.89 \\
& best tuned recipe & +21.76 & $-$0.42 & +21.34 & & +7.90 & \textbf{$-$0.32} & +7.58 \\
& with an estimated corruption model & \multicolumn{3}{c}{unchanged network} & & +10.83 & $-$0.83 & +10.00 \\
& with the true corruption model & \multicolumn{3}{c}{unchanged network} & & +11.00 & $-$0.77 & +10.23 \\
\midrule
\multirow{3}{*}{\shortstack[l]{Fine-tuning,\\no clean-loss\\bound}}
& best of 8 methods & +31.09 & $-$1.98 & +29.11 & & +8.23 & $-$0.34 & +7.89 \\
& best tuned recipe & +33.36 & $-$2.23 & +31.13 & & +7.90 & $-$0.32 & +7.58 \\
& with the true corruption model & +35.75 & $-$2.78 & +32.97 & & +12.39 & $-$0.90 & +11.49 \\
\midrule
\multicolumn{2}{@{}l}{Best of 11 fixed references, kept} & & & +16.48 & & & & +8.25 \\
\bottomrule
\end{tabular}
\caption{\textbf{Fine-tuning on the ViT-Lite under matched data, with and without the clean-accuracy retention bound.} ViT-Lite-7/4 on CIFAR-10-C at severity 5, accuracy changes in points on the held-out test sets. Every method sees the same 64 labeled corrupted and 64 labeled clean images and receives 8 checks. ``Kept by the same rule'' is the checked configuration with the best check score among those losing at most 0.5\% of clean check images, and ``unchanged network'' means that no configuration met the bound. Best of 8 methods: LoRA r4 (Gaussian) and full fine-tuning with L2-SP (contrast), and without the bound LoRA r16 and L2-SP. Tuned recipes add warmup, augmentation and minibatches. The corruption-model rows add images synthesized from a corruption model estimated from the 64 images, which adds no information, or from the true one, which adds the process that generates the condition. Bold: the best value of each column among the methods kept by the rule.}
\label{tab:examples}
\end{table*}

%% file: tables/supp_main_dev.tex
\begin{table*}[t]
\centering
\small
\setlength{\tabcolsep}{4.5pt}
\begin{tabular}{@{}llccccccr@{}}
\toprule
& & \multicolumn{3}{c}{Weight program (\method)} & \multicolumn{2}{c}{Fine-tuning, best of 8} & Fixed refs. & Weights \\
\cmidrule(lr){3-5}\cmidrule(lr){6-7}\cmidrule(lr){8-8}
Network & Condition & gain & clean & net gain & kept & no bound & kept & changed \\
\midrule
\multicolumn{9}{@{}l}{\emph{ImageNet, 1,000 classes, corruptions of ImageNet-C at severity 3 (noise) and 4 (contrast)}} \\
ResNet-18 & Gaussian noise & +10.23 & $-$0.10 & \textbf{+10.13} & +8.70 & -- & +8.00 & 2,152,928 \\
ResNet-18 & contrast reduction & +9.68 & $-$1.13 & \textbf{+8.55} & +3.40 & -- & +5.33 & 33,176 \\
\midrule
\multicolumn{9}{@{}l}{\emph{CIFAR-10, corruptions of CIFAR-10-C at severity 5}} \\
ViT-Lite & Gaussian noise & +31.62 & +0.12 & \textbf{+31.75} & +22.90 & +30.50 ($-$1.43) & +16.08 & 3,846 \\
ViT-Lite & contrast reduction & +17.93 & +0.05 & \textbf{+17.98} & +8.65 & +8.65 (0.00) & +9.48 & 16,648 \\
ResNet-18 & Gaussian noise & +34.98 & $-$0.38 & \textbf{+34.60} & +9.15 & +29.25 ($-$2.38) & +9.10 & 1,718,057 \\
ResNet-18 & contrast reduction & +21.48 & $-$0.33 & \textbf{+21.15} & +9.78 & +26.13 ($-$2.80) & +13.45 & 37,192 \\
\bottomrule
\end{tabular}
\caption{\textbf{\Cref{tab:main} on the selection split.} Accuracy changes in points on the selection split. ImageNet: torchvision's ResNet-18 with its 1,000 classes (\cref{sec:supp_imagenet}). CIFAR-10: ViT-Lite-7/4 trained on CIFAR-10 and an ImageNet-pretrained ResNet-18 fine-tuned on CIFAR-10. Net gain is the gain on the new condition plus the clean change. Every method receives the same 64 labeled images of the condition and 64 labeled clean images (the fine-tuning variant without clean images fits only the 64 of the condition) and the same 8 checks. Kept: the checked configuration with the best check score among those losing at most 0.5\% of clean check images. No bound: the configuration with the best check score at any clean loss, with its clean change in brackets (not run on ImageNet). Fixed refs.: the best of eleven label fits and label-free rules, kept by the same rule. Weights: entries the program changes. Each program row is the first session on its network; every ImageNet session is in \cref{tab:imagenet} and every ViT-Lite session in \cref{tab:sessions}. Bold: the highest net gain among the methods kept by the rule.}
\label{tab:main_dev}
\end{table*}

%% file: tables/supp_examples_dev.tex
\begin{table*}[t]
\centering
\small
\setlength{\tabcolsep}{5pt}
\begin{tabular}{@{}llccccccc@{}}
\toprule
& & \multicolumn{3}{c}{Gaussian noise} & & \multicolumn{3}{c}{Contrast reduction} \\
\cmidrule{3-5}\cmidrule{7-9}
& & gain & clean & net gain & & gain & clean & net gain \\
\midrule
\multicolumn{2}{@{}l}{Weight program (\method)} & \textbf{+31.62} & \textbf{+0.12} & \textbf{+31.75} & & \textbf{+17.93} & +0.05 & \textbf{+17.98} \\
\midrule
\multirow{4}{*}{\shortstack[l]{Fine-tuning,\\kept by the\\same rule}}
& best of 8 methods & +23.52 & $-$0.62 & +22.90 & & +8.65 & +0.00 & +8.65 \\
& best tuned recipe & +22.32 & $-$0.43 & +21.90 & & +8.62 & \textbf{+0.07} & +8.70 \\
& with an estimated corruption model & \multicolumn{3}{c}{unchanged network} & & +11.58 & $-$0.18 & +11.40 \\
& with the true corruption model & \multicolumn{3}{c}{unchanged network} & & +11.85 & $-$0.15 & +11.70 \\
\midrule
\multirow{3}{*}{\shortstack[l]{Fine-tuning,\\no clean-loss\\bound}}
& best of 8 methods & +31.92 & $-$1.43 & +30.50 & & +8.65 & +0.00 & +8.65 \\
& best tuned recipe & +34.10 & $-$1.57 & +32.52 & & +8.62 & +0.07 & +8.70 \\
& with the true corruption model & +35.00 & $-$2.43 & +32.58 & & +13.30 & $-$0.18 & +13.12 \\
\midrule
\multicolumn{2}{@{}l}{Best of 11 fixed references, kept} & & & +16.08 & & & & +9.48 \\
\bottomrule
\end{tabular}
\caption{\textbf{\Cref{tab:examples} on the selection split.} ViT-Lite-7/4 on CIFAR-10-C at severity 5, accuracy changes in points on the selection split. Every method sees the same 64 labeled corrupted and 64 labeled clean images and receives 8 checks. ``Kept by the same rule'' is the checked configuration with the best check score among those losing at most 0.5\% of clean check images, and ``unchanged network'' means that no configuration met the bound. Best of 8 methods: LoRA r4 (Gaussian) and full fine-tuning with L2-SP (contrast), and without the bound LoRA r16 and L2-SP. Tuned recipes add warmup, augmentation and minibatches. The corruption-model rows add images synthesized from a corruption model estimated from the 64 images, which adds no information, or from the true one, which adds the process that generates the condition. Bold: the best value of each column among the methods kept by the rule.}
\label{tab:examples_dev}
\end{table*}

%% file: fig/fig_f2.tex
\begin{figure}[t]
\centering
\includegraphics[width=\linewidth]{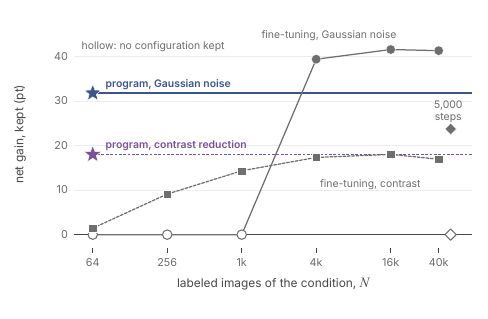}
\caption{\textbf{Under Gaussian noise on the ViT-Lite, full fine-tuning needs substantially more labeled images to match the program under the same retention rule.}
Net gain (points) of full fine-tuning on the ViT-Lite with $N$ labeled images of the condition and $N$ labeled clean images, kept by the same rule as the program (hollow: no configuration kept, so the network is unchanged), against the programs written from 64 images of the condition (stars and lines). Diamonds: 5,000 steps instead of at most 1,000, at $N = 40{,}000$.
Selection split.}
\label{fig:f2}
\end{figure}

%% file: fig/fig_candidates.tex
\begin{figure*}[t]
\centering
\includegraphics[width=\linewidth]{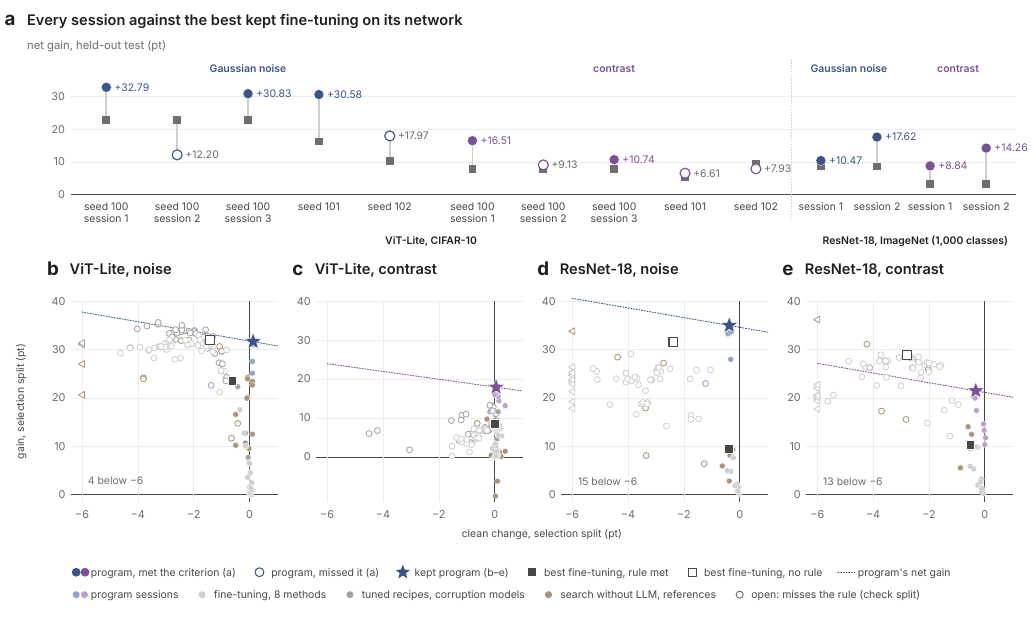}
\caption{\textbf{Program gains and fine-tuning trade-offs across sessions and retention constraints.}
(a)~Net gain of every program session against the best kept fine-tuning on the same network (held-out test sets).
(b--e)~Gain against clean change for every candidate of every method on the four CIFAR-10 instances (selection split); star, the kept program; squares, the best fine-tuning with (filled) and without (open) the retention rule.}
\label{fig:candidates}
\end{figure*}

%% file: tables/tab_sessions.tex
\begin{table*}[t]
\centering
\footnotesize
\setlength{\tabcolsep}{4pt}
\begin{tabular}{@{}lllcccccccccc@{}}
\toprule
& & & \multicolumn{5}{c}{On its own network} & & & \multicolumn{3}{c}{Rerun on the five other networks} \\
\cmidrule(lr){4-8}\cmidrule(lr){11-13}
Condition & Network & Session & gain & clean & net & FT kept & criterion & checks & US\$ & share & clean $\ge -1$ & copied \\
\midrule
Gaussian & seed 100 & 1 & +32.68 & +0.11 & +32.79 & +22.78 & met & 6 & 9.92 & 0.87 & 5 of 5 & $-$0.10\textsuperscript{\dag} \\
 & seed 100 & 2 & +12.73 & $-$0.53 & +12.20 & +22.78 & missed & 8 & 8.36 & 0.98 & 5 of 5 & +0.01\textsuperscript{\dag} \\
 & seed 100 & 3 & +30.80 & +0.03 & +30.83 & +22.78 & met & 8 & 8.75 & 0.91 & 5 of 5 & $-$0.16\textsuperscript{\dag} \\
 & seed 101 &  & +30.85 & $-$0.27 & +30.58 & +16.31 & met & 5 & 7.81 & 0.85 & 5 of 5 & $-$0.34\textsuperscript{\dag} \\
 & seed 102 &  & +18.35 & $-$0.38 & +17.97 & +10.32 & missed & 6 & 8.74 & 0.98 & 5 of 5 & $-$0.45\textsuperscript{\dag} \\
\midrule
contrast & seed 100 & 1 & +17.45 & $-$0.94 & +16.51 & +7.89 & met & 8 & 9.63 & 0.92 & 3 of 5 & $-$0.56\textsuperscript{\dag} \\
 & seed 100 & 2 & +9.51 & $-$0.38 & +9.13 & +7.89 & missed & 6 & 8.90 & 0.96 & 5 of 5 & $-$0.31\textsuperscript{\dag} \\
 & seed 100 & 3 & +11.39 & $-$0.65 & +10.74 & +7.89 & met & 8 & 13.08 & 0.83 & 5 of 5 & $-$1.42\textsuperscript{\dag} \\
 & seed 101 &  & +6.77 & $-$0.16 & +6.61 & +5.43 & missed & 8 & 10.33 & 1.10 & 5 of 5 & $-$0.24\textsuperscript{\dag} \\
 & seed 102 &  & +8.45 & $-$0.52 & +7.93 & +9.52 & missed & 5 & 7.12 & 0.84 & 5 of 5 & $-$0.06\textsuperscript{\dag} \\
\bottomrule
\end{tabular}
\caption{\textbf{Every editing session on the ViT-Lite.} Each row is one LLM session under the protocol of \cref{sec:method} (8 checks, at most 120 minutes), on the seed-100 network or on networks trained from seeds 101 and 102; session 1 on the seed-100 network is the program of \cref{tab:main}. Gain, clean and net gain in points on the held-out test sets. FT kept: the best fine-tuning configuration kept by the same rule on the same network. Criterion: the prespecified performance criterion of the comparative evaluation, fixed before the sessions (gain at least 10, clean change at least $-$1.0, net gain one point above the strongest reference within the program limits). Checks used and API cost per session. Rerun: the session's program run unchanged on the five other networks of its condition. Share: mean gain there over the gain on its own network; clean $\ge -1$: networks whose clean change stays at or above $-$1.0; copied: mean gain when the program's edit is added unchanged instead (\dag: selection split).}
\label{tab:sessions}
\end{table*}

%% file: tables/tab_imagenet.tex
\begin{table}[t]
\centering
\footnotesize
\setlength{\tabcolsep}{3pt}
\begin{tabular}{@{}lccccc@{}}
\toprule
& \multicolumn{3}{c}{Test set} & Selection & ImageNetV2 \\
\cmidrule(lr){2-4}\cmidrule(lr){5-5}\cmidrule(lr){6-6}
& gain & clean & net gain & net gain & clean \\
\midrule
\multicolumn{6}{@{}l}{\emph{Gaussian noise, severity 3; criterion: gain $\ge$ 7.0, net gain $\ge$ +9.84}} \\
Program, session 1 & +10.72 & $-$0.25 & +10.47 & +10.13 & $-$0.37 \\
Program, session 2 & +18.02 & $-$0.40 & +17.62 & +16.38 & $-$0.45 \\
Fine-tuning, kept & +9.45 & $-$0.78 & +8.67 & +8.70 & $-$0.77 \\
Best fixed reference & +9.44 & $-$0.60 & +8.84 & +8.00 & -- \\
BN adaptation & +1.22 & $-$0.14 & +1.08 & +0.93 & $-$0.33 \\
\midrule
\multicolumn{6}{@{}l}{\emph{Contrast reduction, severity 4; criterion: gain $\ge$ 8.5, net gain $\ge$ +7.00}} \\
Program, session 1 & +9.31 & $-$0.47 & +8.84 & +8.55\textsuperscript{*} & $-$0.89 \\
Program, session 2 & +14.48 & $-$0.22 & +14.26 & +15.83 & $-$0.34 \\
Fine-tuning, kept & +3.42 & $-$0.04 & +3.38 & +3.40 & $-$0.27 \\
Best fixed reference & +6.18 & $-$0.18 & +6.00 & +5.33 & -- \\
\bottomrule
\end{tabular}
\caption{\textbf{On ImageNet, all four programs meet the success criterion and exceed fine-tuning.} The ResNet-18 of torchvision with its 1,000 classes, two sessions per condition. Accuracy changes in points. Test set: 38,000 held-out ImageNet validation images, read once; the paired 95\% intervals of the programs' values are within $\pm$0.49 points. Selection: 4,000 images. Criterion on the test set: gain at least the bound shown, clean change at least $-$1.0 and net gain at least one point above the best fixed reference on the same images. \textsuperscript{*}On the selection split this program's clean change is $-$1.13, past the $-$1.0 bound. Fine-tuning, kept: the best of the eight methods on the selection split, BatchNorm only under noise and full under contrast. BN adaptation re-estimates the BatchNorm statistics from the 64 condition images without labels; under contrast it keeps the original network. ImageNetV2: change in top-1 accuracy on the 10,000 ImageNetV2 matched-frequency images (original network 57.28\%), paired 95\% intervals within $\pm$0.43 points.}
\label{tab:imagenet}
\end{table}

%% file: fig/fig_ladder.tex
\begin{figure}[t]
\centering
\includegraphics[width=\linewidth]{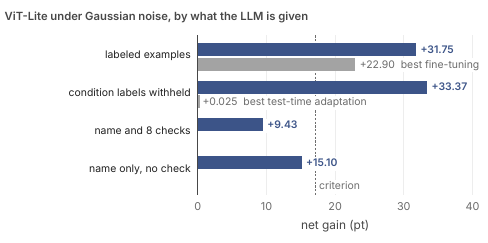}
\caption{\textbf{A program can be written from several forms of information.}
Net gain of the program on the ViT-Lite under Gaussian noise for each form of what the LLM is given (blue), and of the best other method given the same information (gray). The dashed line is the criterion fixed before these sessions. Condition labels withheld: the LLM sees predictions in place of the labels of the corrupted images, which are selected as in the labeled setting; the checks are labeled. One session per level; selection split.}
\label{fig:ladder}
\end{figure}

%% file: tables/tab_controls.tex
\begin{table}[t]
\centering
\footnotesize
\setlength{\tabcolsep}{3pt}
\begin{tabular}{@{}lcc@{}}
\toprule
& Gaussian & contrast \\
\midrule
Weight program (\method) & +31.75 & +17.98 \\
\midrule
\multicolumn{3}{@{}l}{\emph{Ablations of the written program}} \\
\quad block-2 gate removed & +10.05 & -- \\
\quad gates held open (clean $-$5.03) & +19.43 & -- \\
\quad two random units of the block (mean of 5) & +31.86 & -- \\
\midrule
\multicolumn{3}{@{}l}{\emph{Alternatives with the same examples, checks and rule}} \\
\quad gate from the plain mean difference & +31.13 & +5.40 \\
\quad unconditional feature alignment & +7.18 & -- \\
\quad gradient on its entries, from the original & +3.60 & +0.72 \\
\quad gradient on its entries, from its values & +33.35 & +18.00 \\
\quad the LLM without weight values & +10.78 & +11.18 \\
\quad generic gated search, no LLM & +24.12 & unchanged \\
\bottomrule
\end{tabular}
\caption{\textbf{What carries the gain.} Net gain in points on the ViT-Lite on the selection split. Interventions modify parts of the kept program to test their causal contributions and are not reselected with checks. Alternative constructions receive the program's examples and 8 checks (6 for the plain-mean gate, which under contrast reduction runs the noise program's gate form unchanged) and are kept by the same rule; ``unchanged'' means that no configuration met the bound. Gradient rows train exactly the 3,846 (Gaussian) or 16,648 (contrast) entries the program changes, with the fine-tuning grid. The LLM row is a session of the same LLM given the parameter names and shapes, the forward code, the examples and the checks, but no weight value or network output. The search is a post hoc family of gated corrections, written after reading the programs. The two groups answer different questions: what carries the installed gain, and whether other constructions reproduce it.}
\label{tab:controls}
\end{table}

%% file: fig/fig_range.tex
\begin{figure*}[t]
\centering
\includegraphics[width=\linewidth]{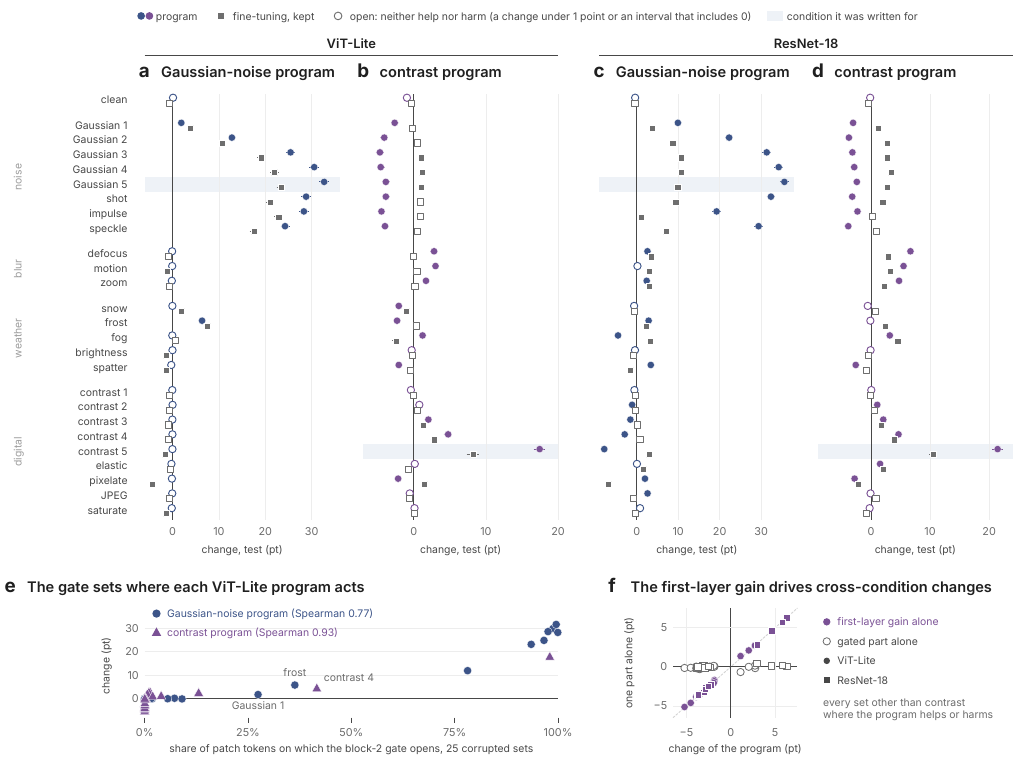}
\caption{\textbf{Installed computations produce distinct target-condition and cross-condition effects.}
(a--d)~Accuracy change against the original network on clean images and on 25 corrupted versions of the CIFAR-10 test set, for the kept program and the kept fine-tuning of the same condition (\cref{tab:main}), with paired 95\% bootstrap intervals: Gaussian noise and contrast at severities 1 to 5 and the other CIFAR-10-C corruptions at severity 5, grouped by family. Filled: help or harm, a change of at least 1 point whose interval excludes 0. Each program was written for severity 5 of its condition (shaded). Held-out test sets.
(e)~Share of patch tokens on which the block-2 gate of each ViT-Lite program opens, against the program's change, on the 25 corrupted sets.
(f)~The first-layer gain determines the cross-condition changes: the two contrast programs split into their always-on first-layer gain and their gated part, each applied alone to the original network, on every set other than contrast reduction where the program helps or harms.
Selection split in (e) and (f).}
\label{fig:range}
\end{figure*}

%% file: tables/tab_reuse.tex
\begin{table}[t]
\centering
\small
\setlength{\tabcolsep}{3.5pt}
\begin{tabular}{@{}llccc@{}}
\toprule
& reused & mean gain & share & clean \\
\midrule
\multirow{3}{*}{Gaussian} & program & \textbf{+28.57} & 0.87 & $-$0.19 to +0.07 \\
& FT recipe & +20.79 & 0.89 & $-$0.85 to $-$0.31 \\
& FT re-selected & +10.44 & 0.45 & $\ge -1.0$ \\
\midrule
\multirow{3}{*}{Contrast} & program & \textbf{+16.04} & 0.92 & $-$1.10 to $-$0.47 \\
& FT recipe & +8.09 & 0.98 & $-$0.55 to $-$0.15 \\
& FT re-selected & +6.57 & 0.80 & $\ge -1.0$ \\
\bottomrule
\end{tabular}
\caption{\textbf{Rerun on five independently trained networks, each program keeps most of its source-network gain.} Programs written on the seed-100 ViT-Lite run with no LLM call on five ViT-Lites trained from other seeds, each with its own 64 labeled examples (Gaussian-noise gains +30.54, +27.55, +27.73, +29.09, +27.92). FT recipe: the best kept fine-tuning configuration on the seed-100 network, applied unchanged; FT re-selected: the full fine-tuning selection repeated on each network. Share: mean gain on the new networks over the gain on the source network. Points on the held-out test sets.}
\label{tab:reuse}
\end{table}

%% file: tables/supp_reuse_dev.tex
\begin{table}[t]
\centering
\small
\setlength{\tabcolsep}{3.5pt}
\begin{tabular}{@{}llccc@{}}
\toprule
& reused & mean gain & share & clean \\
\midrule
\multirow{3}{*}{Gaussian} & program & \textbf{+28.14} & 0.89 & $-$0.07 to 0.00 \\
& FT recipe & +20.71 & 0.88 & $-$0.95 to $-$0.50 \\
& FT re-selected & +10.33 & 0.44 & $\ge -1.0$ \\
\midrule
\multirow{3}{*}{Contrast} & program & \textbf{+17.14} & 0.96 & $-$1.20 to $-$0.65 \\
& FT recipe & +8.91 & 1.03 & $-$0.47 to +0.18 \\
& FT re-selected & +7.19 & 0.83 & $\ge -1.0$ \\
\bottomrule
\end{tabular}
\caption{\textbf{\Cref{tab:reuse} on the selection split.} Programs written on the seed-100 ViT-Lite run with no LLM call on five ViT-Lites trained from other seeds, each with its own 64 labeled examples (Gaussian-noise gains +30.03, +27.18, +27.50, +28.55, +27.43). FT recipe: the best kept fine-tuning configuration on the seed-100 network, applied unchanged; FT re-selected: the full fine-tuning selection repeated on each network. Share: mean gain on the new networks over the gain on the source network. Points on the selection split.}
\label{tab:reuse_dev}
\end{table}

%% file: fig/fig_reuse_r18.tex
\begin{figure}[t]
\centering
\includegraphics[width=\linewidth]{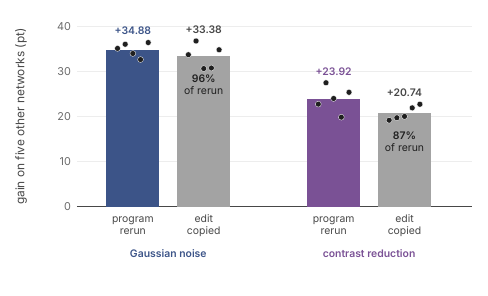}
\caption{\textbf{On ResNet-18s fine-tuned from the same ImageNet weights, rerunning a program gains more than copying its edit.}
Gain (points) of the two ResNet-18 programs rerun on, and of their edits copied to, five ResNet-18s fine-tuned from the same ImageNet weights (dots, one per network; bars, means; percentages, the copied edit's share of the rerun gain).
Selection split.}
\label{fig:reuse_r18}
\end{figure}

%% file: tables/tab_rounds.tex
\begin{table}[t]
\centering
\footnotesize
\setlength{\tabcolsep}{3.5pt}
\begin{tabular}{@{}llcccc@{}}
\toprule
Order & Edit log & added & earlier & clean & met \\
\midrule
Gaussian first & none (first) & +9.12 & $-$0.17 & $-$0.25 & missed \\
 & none & +12.42 & $-$0.02 & $-$0.02 & met \\
 & none & +14.50 & +0.01 & $-$0.21 & met \\
 & log & +5.23 & $-$0.09 & +0.15 & missed \\
 & log & +12.80 & +0.23 & $-$1.47 & met \\
 & log & +13.42 & $-$0.01 & $-$0.80 & met \\
\midrule
contrast first & log & +37.65\textsuperscript{\dag} & $-$0.05\textsuperscript{\dag} & +0.15\textsuperscript{\dag} & met \\
 & none & +35.98\textsuperscript{\dag} & $-$0.28\textsuperscript{\dag} & +0.05\textsuperscript{\dag} & met \\
 & log & +35.90\textsuperscript{\dag} & $-$0.20\textsuperscript{\dag} & +0.03\textsuperscript{\dag} & met \\
 & none & +30.93\textsuperscript{\dag} & $-$0.10\textsuperscript{\dag} & +0.08\textsuperscript{\dag} & met \\
 & log & +34.00\textsuperscript{\dag} & +0.03\textsuperscript{\dag} & +0.00\textsuperscript{\dag} & met \\
 & none & +37.70\textsuperscript{\dag} & $-$0.25\textsuperscript{\dag} & +0.08\textsuperscript{\dag} & met \\
\bottomrule
\end{tabular}
\caption{\textbf{Every second-round session on the ViT-Lite.} Round 1 is the kept program of \cref{tab:main} for the first condition; each row is one LLM session that writes round 2 for the second condition on the network round 1 produced. Added: gain on the second condition from round 1 to round 2; earlier: change on the first condition over the same round; clean: clean change from the original network. Edit log: the session also receives the list of entries round 1 changed. Met: all three criteria fixed before the sessions hold, added at least 10, earlier at least $-$3 and clean at least $-$2.0. Points on the held-out test sets; \dag: selection split.}
\label{tab:rounds}
\end{table}

%% file: tables/tab_stack.tex
\begin{table}[t]
\centering
\small
\setlength{\tabcolsep}{4.5pt}
\begin{tabular}{@{}lcccc@{}}
\toprule
Stream score after round & 1 & 2 & 3 & 4 \\
& {\footnotesize Gauss.} & {\footnotesize defocus} & {\footnotesize snow} & {\footnotesize contrast} \\
\midrule
Programs, one per round & \textbf{2.34} & \textbf{8.22} & \textbf{8.64} & \textbf{8.95} \\
Best fine-tuning stream & 1.48 & 3.63 & 4.56 & 5.15 \\
\bottomrule
\end{tabular}
\caption{\textbf{Exploratory four-round capability updates on ResNet-20, including operations outside the main program protocol (\cref{sec:supp_fd}).} ResNet-20 receives a new corruption each round with 64 labeled examples of it, and each round's program is written for the network the earlier rounds produced. Stream score: mean gain over all four corruptions plus the clean change, in points on the held-out test sets. After round 4 the programs' mean gain is +10.61 at clean $-$1.66, and every corruption ends at or above its gain after its own round. Fine-tuning streams re-select among 8 configurations each round under the same retention rule.}
\label{tab:stack}
\end{table}

%% file: tables/supp_stack_dev.tex
\begin{table}[t]
\centering
\small
\setlength{\tabcolsep}{4.5pt}
\begin{tabular}{@{}lcccc@{}}
\toprule
Stream score after round & 1 & 2 & 3 & 4 \\
& {\footnotesize Gauss.} & {\footnotesize defocus} & {\footnotesize snow} & {\footnotesize contrast} \\
\midrule
Programs, one per round & \textbf{2.31} & \textbf{8.06} & \textbf{8.90} & \textbf{8.97} \\
Best fine-tuning stream & 1.18 & 3.54 & 4.35 & 4.97 \\
\bottomrule
\end{tabular}
\caption{\textbf{Selection-split results for the exploratory four-round ResNet-20 stream (\cref{tab:stack}).} ResNet-20 receives a new corruption each round with 64 labeled examples of it, and each round's program is written for the network the earlier rounds produced. Stream score: mean gain over all four corruptions plus the clean change, in points on the selection split. After round 4 the programs' mean gain is +10.22 at clean $-$1.25, and every corruption ends at or above its gain after its own round. Fine-tuning streams re-select among 8 configurations each round under the same retention rule.}
\label{tab:stack_dev}
\end{table}

%% file: fig/fig_stream.tex
\begin{figure}[t]
\centering
\includegraphics[width=\linewidth]{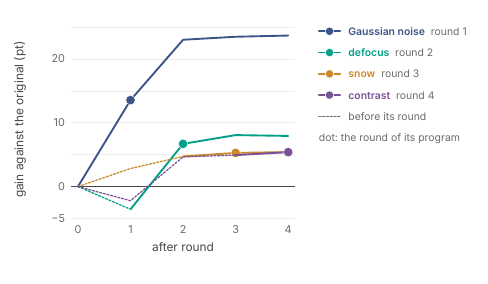}
\caption{\textbf{Gain under each corruption in the ResNet-20 stream of \cref{sec:supp_fd}.} One program per round, each written for the network the earlier rounds produced. Gain against the original network after each round, dashed before the corruption's own round; dots, the round of its program. Exploratory stream that includes sessions outside the main program protocol (\cref{sec:supp_fd}). Held-out test sets.}
\label{fig:stream}
\end{figure}

%% file: references.bib
@inproceedings{hu2021lora,
  title = {{LoRA: Low-Rank Adaptation of Large Language Models}},
  author = {Edward J. Hu and Yelong Shen and Phillip Wallis and Zeyuan Allen-Zhu and Yuanzhi Li and Shean Wang and Lu Wang and Weizhu Chen},
  booktitle = {International Conference on Learning Representations (ICLR)},
  year = {2022},
  eprint = {2106.09685},
  archivePrefix = {arXiv}
}

@inproceedings{houlsby2019adapters,
  title = {{Parameter-Efficient Transfer Learning for NLP}},
  author = {Neil Houlsby and Andrei Giurgiu and Stanislaw Jastrzebski and Bruna Morrone and Quentin de Laroussilhe and Andrea Gesmundo and Mona Attariyan and Sylvain Gelly},
  booktitle = {International Conference on Machine Learning (ICML)},
  year = {2019},
  eprint = {1902.00751},
  archivePrefix = {arXiv}
}

@inproceedings{meng2022rome,
  title = {{Locating and Editing Factual Associations in GPT}},
  author = {Kevin Meng and David Bau and Alex Andonian and Yonatan Belinkov},
  booktitle = {Advances in Neural Information Processing Systems (NeurIPS)},
  year = {2022},
  eprint = {2202.05262},
  archivePrefix = {arXiv}
}

@inproceedings{meng2022memit,
  title = {{Mass-Editing Memory in a Transformer}},
  author = {Kevin Meng and Arnab Sen Sharma and Alex Andonian and Yonatan Belinkov and David Bau},
  booktitle = {International Conference on Learning Representations (ICLR)},
  year = {2023},
  eprint = {2210.07229},
  archivePrefix = {arXiv}
}

@inproceedings{mitchell2021mend,
  title = {{Fast Model Editing at Scale}},
  author = {Eric Mitchell and Charles Lin and Antoine Bosselut and Chelsea Finn and Christopher D. Manning},
  booktitle = {International Conference on Learning Representations (ICLR)},
  year = {2022},
  eprint = {2110.11309},
  archivePrefix = {arXiv}
}

@inproceedings{sinitsin2020editable,
  title = {{Editable Neural Networks}},
  author = {Anton Sinitsin and Vsevolod Plokhotnyuk and Dmitriy Pyrkin and Sergei Popov and Artem Babenko},
  booktitle = {International Conference on Learning Representations (ICLR)},
  year = {2020},
  eprint = {2004.00345},
  archivePrefix = {arXiv}
}

@inproceedings{santurkar2021editing,
  title = {{Editing a classifier by rewriting its prediction rules}},
  author = {Shibani Santurkar and Dimitris Tsipras and Mahalaxmi Elango and David Bau and Antonio Torralba and Aleksander Madry},
  booktitle = {Advances in Neural Information Processing Systems (NeurIPS)},
  year = {2021},
  eprint = {2112.01008},
  archivePrefix = {arXiv}
}

@inproceedings{wang2020tent,
  title = {{Tent: Fully Test-time Adaptation by Entropy Minimization}},
  author = {Dequan Wang and Evan Shelhamer and Shaoteng Liu and Bruno Olshausen and Trevor Darrell},
  booktitle = {International Conference on Learning Representations (ICLR)},
  year = {2021},
  eprint = {2006.10726},
  archivePrefix = {arXiv}
}

@inproceedings{schneider2020bn,
  title = {{Improving robustness against common corruptions by covariate shift adaptation}},
  author = {Steffen Schneider and Evgenia Rusak and Luisa Eck and Oliver Bringmann and Wieland Brendel and Matthias Bethge},
  booktitle = {Advances in Neural Information Processing Systems (NeurIPS)},
  year = {2020},
  eprint = {2006.16971},
  archivePrefix = {arXiv}
}

@inproceedings{sun2019ttt,
  title = {{Test-Time Training with Self-Supervision for Generalization under Distribution Shifts}},
  author = {Yu Sun and Xiaolong Wang and Zhuang Liu and John Miller and Alexei A. Efros and Moritz Hardt},
  booktitle = {International Conference on Machine Learning (ICML)},
  year = {2020},
  eprint = {1909.13231},
  archivePrefix = {arXiv}
}

@inproceedings{niu2023sar,
  title = {{Towards Stable Test-Time Adaptation in Dynamic Wild World}},
  author = {Shuaicheng Niu and Jiaxiang Wu and Yifan Zhang and Zhiquan Wen and Yaofo Chen and Peilin Zhao and Mingkui Tan},
  booktitle = {International Conference on Learning Representations (ICLR)},
  year = {2023},
  eprint = {2302.12400},
  archivePrefix = {arXiv}
}

@inproceedings{hendrycks2019cifarc,
  title = {{Benchmarking Neural Network Robustness to Common Corruptions and Perturbations}},
  author = {Dan Hendrycks and Thomas Dietterich},
  booktitle = {International Conference on Learning Representations (ICLR)},
  year = {2019},
  eprint = {1903.12261},
  archivePrefix = {arXiv}
}

@inproceedings{dosovitskiy2020vit,
  title = {{An Image is Worth 16x16 Words: Transformers for Image Recognition at Scale}},
  author = {Alexey Dosovitskiy and Lucas Beyer and Alexander Kolesnikov and Dirk Weissenborn and Xiaohua Zhai and Thomas Unterthiner and Mostafa Dehghani and Matthias Minderer and Georg Heigold and Sylvain Gelly and Jakob Uszkoreit and Neil Houlsby},
  booktitle = {International Conference on Learning Representations (ICLR)},
  year = {2021},
  eprint = {2010.11929},
  archivePrefix = {arXiv}
}

@article{hassani2021cct,
  title = {{Escaping the Big Data Paradigm with Compact Transformers}},
  author = {Ali Hassani and Steven Walton and Nikhil Shah and Abulikemu Abuduweili and Jiachen Li and Humphrey Shi},
  journal = {arXiv preprint arXiv:2104.05704},
  year = {2021}
}

@inproceedings{li2018l2sp,
  title = {{Explicit Inductive Bias for Transfer Learning with Convolutional Networks}},
  author = {Xuhong Li and Yves Grandvalet and Franck Davoine},
  booktitle = {International Conference on Machine Learning (ICML)},
  year = {2018},
  eprint = {1802.01483},
  archivePrefix = {arXiv}
}

@article{chaudhry2019tiny,
  title = {{On Tiny Episodic Memories in Continual Learning}},
  author = {Arslan Chaudhry and Marcus Rohrbach and Mohamed Elhoseiny and Thalaiyasingam Ajanthan and Puneet K. Dokania and Philip H. S. Torr and Marc'Aurelio Ranzato},
  journal = {arXiv preprint arXiv:1902.10486},
  year = {2019}
}

@inproceedings{ilharco2022taskarith,
  title = {{Editing Models with Task Arithmetic}},
  author = {Gabriel Ilharco and Marco Tulio Ribeiro and Mitchell Wortsman and Suchin Gururangan and Ludwig Schmidt and Hannaneh Hajishirzi and Ali Farhadi},
  booktitle = {International Conference on Learning Representations (ICLR)},
  year = {2023},
  eprint = {2212.04089},
  archivePrefix = {arXiv}
}

@article{novikov2025alphaevolve,
  title = {{AlphaEvolve: A coding agent for scientific and algorithmic discovery}},
  author = {Alexander Novikov and Ng{\^a}n V{\~u} and Marvin Eisenberger and Emilien Dupont and Po-Sen Huang and Adam Zsolt Wagner and Sergey Shirobokov and Borislav Kozlovskii and Francisco J. R. Ruiz and Abbas Mehrabian and M. Pawan Kumar and Abigail See and Swarat Chaudhuri and George Holland and Alex Davies and Sebastian Nowozin and Pushmeet Kohli and Matej Balog},
  journal = {arXiv preprint arXiv:2506.13131},
  year = {2025}
}

@inproceedings{ma2023eureka,
  title = {{Eureka: Human-Level Reward Design via Coding Large Language Models}},
  author = {Yecheng Jason Ma and William Liang and Guanzhi Wang and De-An Huang and Osbert Bastani and Dinesh Jayaraman and Yuke Zhu and Linxi Fan and Anima Anandkumar},
  booktitle = {International Conference on Learning Representations (ICLR)},
  year = {2024},
  eprint = {2310.12931},
  archivePrefix = {arXiv}
}

@article{wang2023voyager,
  title = {{Voyager: An Open-Ended Embodied Agent with Large Language Models}},
  author = {Guanzhi Wang and Yuqi Xie and Yunfan Jiang and Ajay Mandlekar and Chaowei Xiao and Yuke Zhu and Linxi Fan and Anima Anandkumar},
  journal = {Transactions on Machine Learning Research (TMLR)},
  year = {2024},
  eprint = {2305.16291},
  archivePrefix = {arXiv}
}

@inproceedings{andrychowicz2016l2l,
  title = {{Learning to learn by gradient descent by gradient descent}},
  author = {Marcin Andrychowicz and Misha Denil and Sergio Gomez and Matthew W. Hoffman and David Pfau and Tom Schaul and Brendan Shillingford and Nando de Freitas},
  booktitle = {Advances in Neural Information Processing Systems (NeurIPS)},
  year = {2016},
  eprint = {1606.04474},
  archivePrefix = {arXiv}
}

@inproceedings{chen2023lion,
  title = {{Symbolic Discovery of Optimization Algorithms}},
  author = {Xiangning Chen and Chen Liang and Da Huang and Esteban Real and Kaiyuan Wang and Yao Liu and Hieu Pham and Xuanyi Dong and Thang Luong and Cho-Jui Hsieh and Yifeng Lu and Quoc V. Le},
  booktitle = {Advances in Neural Information Processing Systems (NeurIPS)},
  year = {2023},
  eprint = {2302.06675},
  archivePrefix = {arXiv}
}

@inproceedings{hartvigsen2023grace,
  title = {{Aging with GRACE: Lifelong Model Editing with Discrete Key-Value Adaptors}},
  author = {Thomas Hartvigsen and Swami Sankaranarayanan and Hamid Palangi and Yoon Kim and Marzyeh Ghassemi},
  booktitle = {Advances in Neural Information Processing Systems (NeurIPS)},
  year = {2023},
  eprint = {2211.11031},
  archivePrefix = {arXiv}
}

@inproceedings{ainsworth2022rebasin,
  title = {{Git Re-Basin: Merging Models modulo Permutation Symmetries}},
  author = {Samuel K. Ainsworth and Jonathan Hayase and Siddhartha Srinivasa},
  booktitle = {International Conference on Learning Representations (ICLR)},
  year = {2023},
  eprint = {2209.04836},
  archivePrefix = {arXiv}
}

@inproceedings{yadav2023ties,
  title = {{TIES-Merging: Resolving Interference When Merging Models}},
  author = {Prateek Yadav and Derek Tam and Leshem Choshen and Colin Raffel and Mohit Bansal},
  booktitle = {Advances in Neural Information Processing Systems (NeurIPS)},
  year = {2023},
  eprint = {2306.01708},
  archivePrefix = {arXiv}
}

@article{li2026weightloop,
  title = {{WeightLoop: Weight-Space Programming with Experimental Feedback}},
  author = {Mengran Li and Bo Li and Weijing Zhao and Chengyang Zhang and Boyu Zhang and Jinfeng Xu and Zelin Zang and Victor C. M. Leung and Stan Z. Li},
  journal = {arXiv preprint arXiv:2611.07319},
  year = {2026}
}

@inproceedings{lu2024discopop,
  title = {{Discovering Preference Optimization Algorithms with and for Large Language Models}},
  author = {Chris Lu and Samuel Holt and Claudio Fanconi and Alex J. Chan and Jakob Foerster and Mihaela van der Schaar and Robert Tjarko Lange},
  booktitle = {Advances in Neural Information Processing Systems (NeurIPS)},
  year = {2024},
  eprint = {2406.08414},
  archivePrefix = {arXiv}
}

@inproceedings{ha2016hypernetworks,
  title = {{HyperNetworks}},
  author = {David Ha and Andrew Dai and Quoc V. Le},
  booktitle = {International Conference on Learning Representations (ICLR)},
  year = {2017},
  eprint = {1609.09106},
  archivePrefix = {arXiv}
}

@inproceedings{navon2023dws,
  title = {{Equivariant Architectures for Learning in Deep Weight Spaces}},
  author = {Aviv Navon and Aviv Shamsian and Idan Achituve and Ethan Fetaya and Gal Chechik and Haggai Maron},
  booktitle = {International Conference on Machine Learning (ICML)},
  year = {2023},
  eprint = {2301.12780},
  archivePrefix = {arXiv}
}

@inproceedings{zhou2023nfn,
  title = {{Permutation Equivariant Neural Functionals}},
  author = {Allan Zhou and Kaien Yang and Kaylee Burns and Adriano Cardace and Yiding Jiang and Samuel Sokota and J. Zico Kolter and Chelsea Finn},
  booktitle = {Advances in Neural Information Processing Systems (NeurIPS)},
  year = {2023},
  eprint = {2302.14040},
  archivePrefix = {arXiv}
}

@inproceedings{schurholt2021hyperrep,
  title = {{Hyper-Representations: Self-Supervised Representation Learning on Neural Network Weights for Model Characteristic Prediction}},
  author = {Konstantin Sch{\"u}rholt and Dimche Kostadinov and Damian Borth},
  booktitle = {Advances in Neural Information Processing Systems (NeurIPS)},
  year = {2021},
  eprint = {2110.15288},
  archivePrefix = {arXiv}
}

@inproceedings{lindner2023tracr,
  title = {{Tracr: Compiled Transformers as a Laboratory for Interpretability}},
  author = {David Lindner and J{\'a}nos Kram{\'a}r and Sebastian Farquhar and Matthew Rahtz and Thomas McGrath and Vladimir Mikulik},
  booktitle = {Advances in Neural Information Processing Systems (NeurIPS)},
  year = {2023},
  eprint = {2301.05062},
  archivePrefix = {arXiv}
}

@inproceedings{weiss2021rasp,
  title = {{Thinking Like Transformers}},
  author = {Gail Weiss and Yoav Goldberg and Eran Yahav},
  booktitle = {International Conference on Machine Learning (ICML)},
  year = {2021},
  eprint = {2106.06981},
  archivePrefix = {arXiv}
}

@inproceedings{friedman2023ltp,
  title = {{Learning Transformer Programs}},
  author = {Dan Friedman and Alexander Wettig and Danqi Chen},
  booktitle = {Advances in Neural Information Processing Systems (NeurIPS)},
  year = {2023},
  eprint = {2306.01128},
  archivePrefix = {arXiv}
}

@inproceedings{lake2018scan,
  title = {{Generalization without systematicity: On the compositional skills of sequence-to-sequence recurrent networks}},
  author = {Brenden M. Lake and Marco Baroni},
  booktitle = {International Conference on Machine Learning (ICML)},
  year = {2018},
  eprint = {1711.00350},
  archivePrefix = {arXiv}
}

@inproceedings{chevalierboisvert2018babyai,
  title = {{BabyAI: A Platform to Study the Sample Efficiency of Grounded Language Learning}},
  author = {Maxime Chevalier-Boisvert and Dzmitry Bahdanau and Salem Lahlou and Lucas Willems and Chitwan Saharia and Thien Huu Nguyen and Yoshua Bengio},
  booktitle = {International Conference on Learning Representations (ICLR)},
  year = {2019},
  eprint = {1810.08272},
  archivePrefix = {arXiv}
}

@inproceedings{he2016resnet,
  title={Deep Residual Learning for Image Recognition},
  DOI={10.1109/cvpr.2016.90},
  booktitle={2016 IEEE Conference on Computer Vision and Pattern Recognition (CVPR)},
  publisher={IEEE},
  author={He, Kaiming and Zhang, Xiangyu and Ren, Shaoqing and Sun, Jian},
  year={2016},
  month=jun,
  pages={770--778} }

@inbook{mccloskey1989catastrophic,
  title={Catastrophic Interference in Connectionist Networks: The Sequential Learning Problem},
  ISBN={9780125433242},
  ISSN={0079-7421},
  DOI={10.1016/s0079-7421(08)60536-8},
  booktitle={Psychology of Learning and Motivation},
  publisher={Elsevier},
  author={McCloskey, Michael and Cohen, Neal J.},
  year={1989},
  pages={109--165} }

@article{french1999catastrophic,
  title={Catastrophic forgetting in connectionist networks},
  volume={3},
  ISSN={1364-6613},
  DOI={10.1016/s1364-6613(99)01294-2},
  number={4},
  journal={Trends in Cognitive Sciences},
  publisher={Elsevier BV},
  author={French, R},
  year={1999},
  month=apr,
  pages={128--135} }

@article{romeraparedes2024funsearch,
  title={Mathematical discoveries from program search with large language models},
  volume={625},
  ISSN={1476-4687},
  DOI={10.1038/s41586-023-06924-6},
  number={7995},
  journal={Nature},
  publisher={Springer Science and Business Media LLC},
  author={Romera-Paredes, Bernardino and Barekatain, Mohammadamin and Novikov, Alexander and Balog, Matej and Kumar, M. Pawan and Dupont, Emilien and Ruiz, Francisco J. R. and Ellenberg, Jordan S. and Wang, Pengming and Fawzi, Omar and Kohli, Pushmeet and Fawzi, Alhussein},
  year={2024},
  pages={468--475} }

@inproceedings{liang2023cap,
  title={Code as Policies: Language Model Programs for Embodied Control},
  DOI={10.1109/icra48891.2023.10160591},
  booktitle={2023 IEEE International Conference on Robotics and Automation (ICRA)},
  publisher={IEEE},
  author={Liang, Jacky and Huang, Wenlong and Xia, Fei and Xu, Peng and Hausman, Karol and Ichter, Brian and Florence, Pete and Zeng, Andy},
  year={2023},
  month=may,
  pages={9493--9500} }

@inbook{bau2020rewriting,
  title={Rewriting a Deep Generative Model},
  ISBN={9783030584528},
  ISSN={1611-3349},
  DOI={10.1007/978-3-030-58452-8_21},
  booktitle={Computer Vision -- ECCV 2020},
  publisher={Springer International Publishing},
  author={Bau, David and Liu, Steven and Wang, Tongzhou and Zhu, Jun-Yan and Torralba, Antonio},
  year={2020},
  pages={351--369} }

@article{li2018lwf,
  title={Learning without Forgetting},
  volume={40},
  ISSN={1939-3539},
  DOI={10.1109/tpami.2017.2773081},
  number={12},
  journal={IEEE Transactions on Pattern Analysis and Machine Intelligence},
  publisher={Institute of Electrical and Electronics Engineers (IEEE)},
  author={Li, Zhizhong and Hoiem, Derek},
  year={2018},
  month=dec,
  pages={2935--2947} }

@article{kirkpatrick2017ewc,
  title={Overcoming catastrophic forgetting in neural networks},
  volume={114},
  ISSN={1091-6490},
  DOI={10.1073/pnas.1611835114},
  number={13},
  journal={Proceedings of the National Academy of Sciences},
  publisher={National Academy of Sciences},
  author={Kirkpatrick, James and Pascanu, Razvan and Rabinowitz, Neil and Veness, Joel and Desjardins, Guillaume and Rusu, Andrei A. and Milan, Kieran and Quan, John and Ramalho, Tiago and Grabska-Barwinska, Agnieszka and Hassabis, Demis and Clopath, Claudia and Kumaran, Dharshan and Hadsell, Raia},
  year={2017},
  month=mar,
  pages={3521--3526} }

@inproceedings{recht2019imagenetv2,
  title = {{Do ImageNet Classifiers Generalize to ImageNet?}},
  author = {Benjamin Recht and Rebecca Roelofs and Ludwig Schmidt and Vaishaal Shankar},
  booktitle = {International Conference on Machine Learning (ICML)},
  series = {Proceedings of Machine Learning Research},
  volume = {97},
  pages = {5389--5400},
  year = {2019}
}

@article{russakovsky2014imagenet,
  title={{ImageNet} Large Scale Visual Recognition Challenge},
  volume={115},
  ISSN={1573-1405},
  DOI={10.1007/s11263-015-0816-y},
  number={3},
  journal={International Journal of Computer Vision},
  publisher={Springer Science and Business Media LLC},
  author={Russakovsky, Olga and Deng, Jia and Su, Hao and Krause, Jonathan and Satheesh, Sanjeev and Ma, Sean and Huang, Zhiheng and Karpathy, Andrej and Khosla, Aditya and Bernstein, Michael and Berg, Alexander C. and Fei-Fei, Li},
  year={2015},
  month=apr,
  pages={211--252} }

@inproceedings{loula2018rearranging,
  title={Rearranging the Familiar: Testing Compositional Generalization in Recurrent Networks},
  DOI={10.18653/v1/w18-5413},
  booktitle={Proceedings of the 2018 EMNLP Workshop BlackboxNLP: Analyzing and Interpreting Neural Networks for NLP},
  publisher={Association for Computational Linguistics},
  author={Loula, Jo{\~a}o and Baroni, Marco and Lake, Brenden},
  year={2018},
  pages={108--114} }

@inproceedings{zhang2021cec,
  title={Few-Shot Incremental Learning with Continually Evolved Classifiers},
  DOI={10.1109/cvpr46437.2021.01227},
  booktitle={2021 IEEE/CVF Conference on Computer Vision and Pattern Recognition (CVPR)},
  publisher={IEEE},
  author={Zhang, Chi and Song, Nan and Lin, Guosheng and Zheng, Yun and Pan, Pan and Xu, Yinghui},
  year={2021},
  month=jun,
  pages={12450--12459} }

@misc{takehi2026sparks,
  title = {{Sparks of Weight-Space Programming: LLMs can natively read \& write neural network weights}},
  author = {Rikiya Takehi and Jayden Teoh and Alex Zhang and Omar Khattab},
  howpublished = {Blog post, \url{https://rikiyat.github.io/blog/weight-space-programming/}},
  note = {Published November 2, 2026},
  year = {2026}
}
